%% file: 0-main.tex
\documentclass{article}

\usepackage{microtype}
\usepackage{graphicx}
\usepackage{subfigure}
\usepackage{booktabs} 

\usepackage{hyperref}

\usepackage[accepted]{icml2018}

\icmltitlerunning{Style Tokens: Unsupervised Style Modeling, Control and Transfer in End-to-End Speech Synthesis}

\usepackage{todonotes}

\begin{document}

\twocolumn[

\icmltitle{Style Tokens: Unsupervised Style Modeling, Control and Transfer in End-to-End Speech Synthesis}

\begin{icmlauthorlist}
\icmlauthor{Yuxuan Wang}{google}
\icmlauthor{Daisy Stanton}{google}
\icmlauthor{Yu Zhang}{google}
\icmlauthor{RJ Skerry-Ryan}{google}
\icmlauthor{Eric Battenberg}{google}
\icmlauthor{Joel Shor}{google}
\icmlauthor{Ying Xiao}{google}
\icmlauthor{Fei Ren}{google}
\icmlauthor{Ye Jia}{google}
\icmlauthor{Rif A. Saurous}{google}
\end{icmlauthorlist}

\icmlaffiliation{google}{Google, Inc.}

\icmlcorrespondingauthor{Yuxuan Wang}{yxwang@google.com}


\vskip 0.3in
]



\printAffiliationsAndNotice{}  

\begin{abstract}
In this work, we propose ``global style tokens'' (GSTs), a bank of embeddings that are jointly trained within Tacotron, a state-of-the-art end-to-end speech synthesis system. The embeddings are trained with no explicit labels, yet learn to model a large range of acoustic expressiveness. GSTs lead to a rich set of significant results. The soft interpretable ``labels'' they generate can be used to control synthesis in novel ways, such as varying speed and speaking style -- independently of the text content. They can also be used for style transfer, replicating the speaking style of a single audio clip across an entire long-form text corpus. When trained on noisy, unlabeled found data, GSTs learn to factorize noise and speaker identity, providing a path towards highly scalable but robust speech synthesis.

\end{abstract}

\input{1-introduction}

\input{2-model}

\input{3-model-details}
\input{4-interpretation}

\input{5-related}

\input{6-experiments-expressive}

\input{7-experiments-noisy}

\input{8-conclusions}

\section*{Acknowledgements}

The authors thank Aren Jansen, Rob Clark, Zhifeng Chen, Ron J. Weiss, Mike Schuster, Yonghui Wu, Patrick Nguyen, and the Machine Hearing, Google Brain and Google TTS teams for their helpful
discussions and feedback.

\bibliography{example_paper,9-references}
\bibliographystyle{icml2018}

\end{document}

%% file: 1-introduction.tex
\section{Introduction}
\label{sec.intro}

The past few years have seen exciting developments in the use of deep neural networks to synthesize natural-sounding human speech \cite{zen2016fast,van2016wavenet,yx2017tacotron,arik2017deep,taigman2017voice,shen2017natural}. As text-to-speech (TTS) models have rapidly improved,
there is a growing opportunity for a number of applications, such as audiobook narration, news readers, and conversational assistants. Neural models show the potential to robustly synthesize expressive long-form speech, and yet research in this area is still in its infancy.

To deliver true human-like speech, a TTS system must learn to model prosody. Prosody is the confluence of a number of phenomena in speech, such as paralinguistic information, intonation, stress, and style. In this work we focus on \textit{style modeling}, the goal of which is to provide models the capability to choose a speaking style appropriate for the given context. While difficult to define precisely, style contains rich information, such as intention and emotion, and influences the speaker's choice of intonation and flow. Proper stylistic rendering affects overall perception (see e.g. ``affective prosody'' in \cite{taylor2009text}), which is important for applications such as audiobooks and newsreaders.

Style modeling presents several challenges. First, there is no objective measure of ``correct'' prosodic style, making both modeling and evaluation difficult. Acquiring annotations for large datasets can be costly and similarly problematic, since human raters often disagree.
Second, the high dynamic range in expressive voices is difficult to model. Many TTS models, including recent end-to-end systems, only learn an averaged prosodic distribution over their input data, generating less expressive speech especially for long-form phrases. Furthermore, they often lack the ability to control the expression with which speech is synthesized.

This work attempts to address the above issues by introducing ``global style tokens'' (GSTs) to Tacotron \cite{yx2017tacotron, shen2017natural}, a state-of-the-art end-to-end TTS model. GSTs are trained without any prosodic labels, and yet uncover a large range of expressive styles. The internal architecture itself produces soft interpretable ``labels'' that can be used to perform various style control and transfer tasks, leading to significant improvements for expressive long-form synthesis. GSTs can be directly applied to noisy, unlabeled found data, providing a path towards highly scalable but robust speech synthesis.

%% file: 2-model.tex
\section{Model Architecture}
\label{sec.model}

\begin{figure*}[t]
\vskip 0.2in
\begin{center}
\centerline{\includegraphics[scale=0.43]{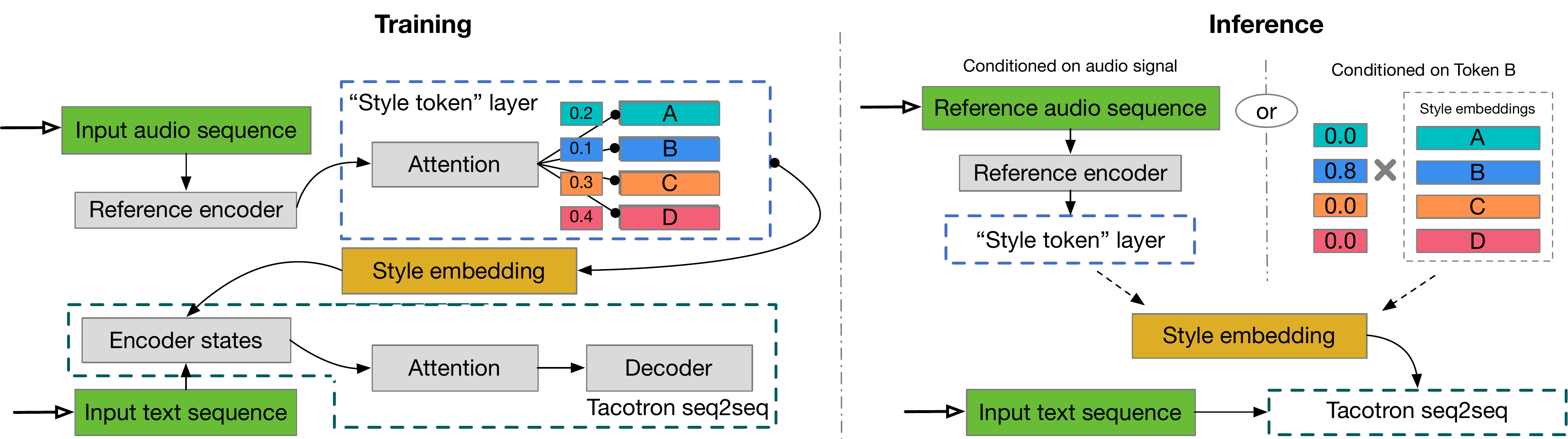}}
\caption{Model diagram. During \textbf{training}, the log-mel spectrogram of the training target is fed to the reference encoder followed by a style token layer. The resulting style embedding is used to condition the Tacotron text encoder states. During \textbf{inference}, we can feed an arbitrary reference signal to synthesize text with its speaking style. Alternatively, we can remove the reference encoder and directly control synthesis using the learned interpretable tokens.}

\label{fig.model}
\end{center}
\vskip -0.2in
\end{figure*}

Our model is based on Tacotron \cite{yx2017tacotron, shen2017natural}, a 
sequence-to-sequence (seq2seq) model that predicts mel spectrograms directly from grapheme or phoneme inputs. These mel spectrograms are converted to waveforms either by a low-resource inversion algorithm \cite{griffin1984signal} or a neural vocoder such as WaveNet \cite{van2016wavenet}. We point out that, for Tacotron, the choice of vocoder does not affect prosody, which is modeled by the seq2seq model.

Our proposed GST model, illustrated in Figure \ref{fig.model},
consists of a reference encoder, style attention, style embedding, and sequence-to-sequence (Tacotron) model.

\subsection{Training}
\label{sec.model.training}
During training, information flows through the model as follows:
\vspace{-0.25cm}

\begin{itemize}
    \item The \textbf{reference encoder}, proposed in \cite{rj2018transfer}, compresses the prosody of a variable-length audio signal into a fixed-length vector, which we call the \textit{reference embedding}. During training, the reference signal is ground-truth audio.
    \vspace{-0.1cm}
    
    \item The reference embedding is passed to a style token layer, where it is used as the query vector to an \textbf{attention module}. Here, attention is not used to learn an alignment. Instead, it learns a similarity measure between the reference embedding and each token in a bank of \textbf{randomly initialized embeddings}. This set of embeddings, which we alternately call \textit{global style tokens}, GSTs, or token embeddings, are shared across all training sequences. 
    \vspace{-0.1cm}

    \item The attention module outputs a set of combination weights that represent the contribution of each style token to the encoded reference embedding. The weighted sum of the GSTs, which we call the \textit{style embedding}, is passed to the text encoder for conditioning at every timestep.
    \vspace{-0.1cm}

    \item The style token layer is jointly trained with the rest of the model, driven only by the reconstruction loss from the Tacotron decoder. GSTs thus do not require any explicit style or prosody labels.

\end{itemize}

\subsection{Inference}
\label{sec.model.inference}
The GST architecture is designed for powerful and flexible control in inference mode. In this mode, information can flow through the model in one of two ways:
\vspace{-0.2cm}
\begin{enumerate}
    \item We can directly condition the text encoder on certain tokens, as depicted on the right-hand side of the inference-mode diagram in Figure \ref{fig.model} (``Conditioned on Token B''). This allows for style control and manipulation without a reference signal.
    \vspace{-0.15cm}    
    \item We can feed a different audio signal (whose transcript does not need to match the text to be synthesized) to achieve style transfer. This is depicted on the left-hand side of the inference-mode diagram in Figure \ref{fig.model} (``Conditioned on audio signal'').
    \vspace{-0.15cm}
\end{enumerate}
\vspace{-0.1cm}
These will be discussed in more detail in Section \ref{sec.expt}.

%% file: 3-model-details.tex
\section{Model Details}

\subsection{Tacotron Architecture}
For our baseline and GST-augmented Tacotron systems, we use the same architecture and hyperparameters as \cite{yx2017tacotron} except for a few details.
We use phoneme inputs to speed up training, and slightly change the decoder, replacing GRU cells with two layers of 256-cell LSTMs; these are regularized using zoneout \cite{krueger2016zoneout} with probability 0.1. The decoder outputs 80-channel log-mel spectrogram energies, two frames at a time, which are run through a dilated convolution network that outputs linear spectrograms. We run these through Griffin-Lim for fast waveform reconstruction. It is straightforward to replace Griffin-Lim by a WaveNet vocoder to improve the audio fidelity \cite{shen2017natural}.

The baseline model achieves a 4.0 mean opinion score (MOS), outperforming the 3.82 MOS reported in \cite{yx2017tacotron} on the same evaluation set. It is thus a very strong baseline.

\subsection{Style Token Architecture}

\subsubsection{Reference Encoder}

The reference encoder is made up of a convolutional stack, followed by an RNN. 
It takes as input a log-mel spectrogram,
which is first passed to a stack of six 2-D convolutional layers with 3$\times$3 kernel, 2$\times$2 stride, batch normalization and ReLU activation function. We use 32, 32, 64, 64, 128 and 128 output channels for the 6 convolutional layers, respectively. The resulting output tensor is then shaped back to 3 dimensions (preserving the output time resolution) and fed to a single-layer 128-unit unidirectional GRU. The last GRU state serves as the reference embedding, which is then fed as input to the style token layer.

\subsubsection{Style Token Layer}

The style token layer is made up of a bank of style token embeddings and an attention module. Unless stated otherwise, our experiments use 10 tokens, which we found sufficient to represent a small but rich variety of prosodic dimensions in the training data.
To match the dimensionality of the text encoder state, each token embedding is 256-D.
Similarly, the text encoder state uses a tanh activation; we found that applying a tanh activation to GSTs before applying attention led to greater token diversity. 
The content-based tanh attention uses a softmax activation to output a set of combination weights over the tokens; the resulting weighted combination of GSTs is then used for conditioning. We experimented with different combinations of conditioning sites, and found that replicating the style embedding and simply adding it to every text encoder state performed the best.

While we use content-based attention as a similarity measure in this work, it is trivial to substitute alternatives. Dot-product attention, location-based attention, or even combinations of attention mechanisms may learn different types of style tokens. In our experiments, we found that using multi-head attention \cite{vaswani2017attention} significantly improves style transfer performance, and, moreover, is more effective than simply increasing the number of tokens. When using $h$ attention heads, we set the token embedding size to be $256 / h$ and concatenate the attention outputs, such that the final style embedding size remains the same.


%% file: 4-interpretation.tex
\section{Model Interpretation}
\label{sec.interp}

\subsection{End-to-End Clustering/Quantization}

Intuitively, the GST model can be thought of as an end-to-end method for decomposing
the reference embedding into a set of basis vectors or soft clusters -- i.e. the style tokens.
As mentioned above, the contribution of each style token is represented by an attention score, but can be replaced with any desired similarity measure.
The GST layer is conceptually somewhat similar to the VQ-VAE encoder \cite{van2017neural}, in that it learns a quantized representation of its input. We also experimented with replacing the GST layer with a discrete, VQ-like lookup table layer, but have not seen comparable results yet.

This decomposition concept can also be generalized to other models, e.g. the factorized variational latent model in \cite{hsu2017unsupervised}, which exploits the multi-scale nature of a speech signal by explicitly formulating it within a factorized hierarchical graphical model. 
Its sequence-dependent priors are formulated by an embedding table, which is similar to GSTs but without the attention-based clustering. GSTs could potentially be used to reduce the required samples to learn each prior embedding.

\subsection{Memory-Augmented Neural Network}
GST embeddings can also be viewed as an external memory that stores style information extracted from training data. The reference signal guides memory writes at training time, and memory reads at inference time. We may leverage recent advances from memory-augmented networks \cite{graves2014neural} to further improve GST learning.


%% file: 5-related.tex
\section{Related Work}

Prosody and speaking style models have been studied for decades in the TTS community. However, most existing models require explicit labels, such as emotion or speaker codes \cite{luong2017adapting}. While a small amount of research has explored automatic labeling, learning is still supervised, requiring expensive annotations for model training. AuToBI, for example, \cite{rosenberg2010autobi} aims to produce ToBI \cite{silverman1992tobi} labels that can be used by other TTS models. However, AuToBI still needs annotations for training, and ToBI, as a hand-designed label system, is known to have limited performance \cite{wightman2002tobi}.

Cluster-based modeling \cite{eyben2012unsupervised, jauk2017unsupervised} is related to our work. Jauk \yrcite{jauk2017unsupervised}, for example, uses $i$-vectors \cite{dehak2011front} and other acoustic features to cluster the training set and train models in different partitions. These methods rely on a complex set of hand-designed features, however, and require training a neutral voice model in a separate step.

As mentioned previously, \cite{rj2018transfer} introduces the reference embedding used in this work, and shows that it can be used to transfer prosody from a reference signal. This embedding does not enable interpretable style control, however, and we show in Section \ref{sec.expt} that it generalizes poorly on some style transfer tasks.

Our work substantially extends the research in
\cite{wang2017uncovering}, but there are several fundamental differences. First, \cite{wang2017uncovering} uses a single frame from the Tacotron \textit{decoder} as the query to learn tokens. It thus only models ``local''  variations that primarily correspond to F0. GSTs instead use a summary of the entire reference signal as input, and are thus able to uncover both local and global attributes that are essential for expressive synthesis.  Second, in contrast to the decoder-side conditioning in \cite{wang2017uncovering}, the design of GSTs allows textual input to be conditioned on a disentangled style embedding. We show crucial implications of this for style control and transfer in Section \ref{sec.expt.transfer}.
Finally, GSTs can be applied to both clean recordings and noisy found data. We discuss this and its significance in detail in Section \ref{sec.expt.found}.

%% file: 6-experiments-expressive.tex
\section{Experiments: Style Control and Transfer}
\label{sec.expt}

In this section, we measure the ability of GSTs to control and transfer speaking style, using the inference methods from Section \ref{sec.model.inference}.

We train models using 147 hours of American English audiobook data. These are read by the 2013 Blizzard Challenge speaker, Catherine Byers,  in an animated and emotive storytelling style. Some books contain very expressive character voices with high dynamic range, which are challenging to model.

As is common for generative models, objective metrics often do not correlate well with perception 
\cite{theis2015note}. While we use visualizations for some experiments below, we strongly 
encourage readers to listen to the samples provided 
on our \href{https://google.github.io/tacotron/publications/global_style_tokens/}{demo page}.

\begin{figure}[t]
\centering
\subfigure[Sentence A]{
\label{fig.tokens.a}
\includegraphics[scale=0.27]{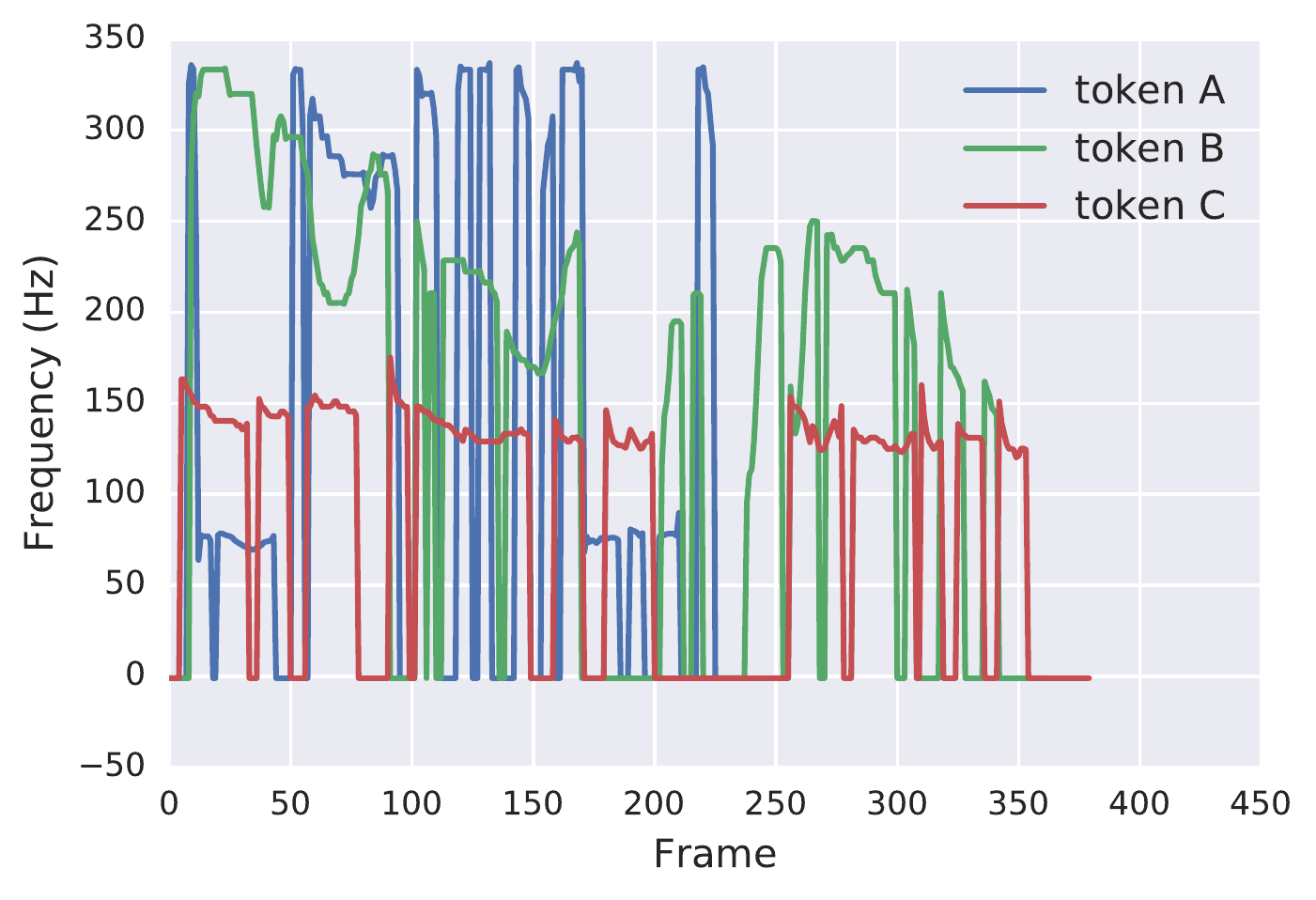}
\includegraphics[scale=0.27]{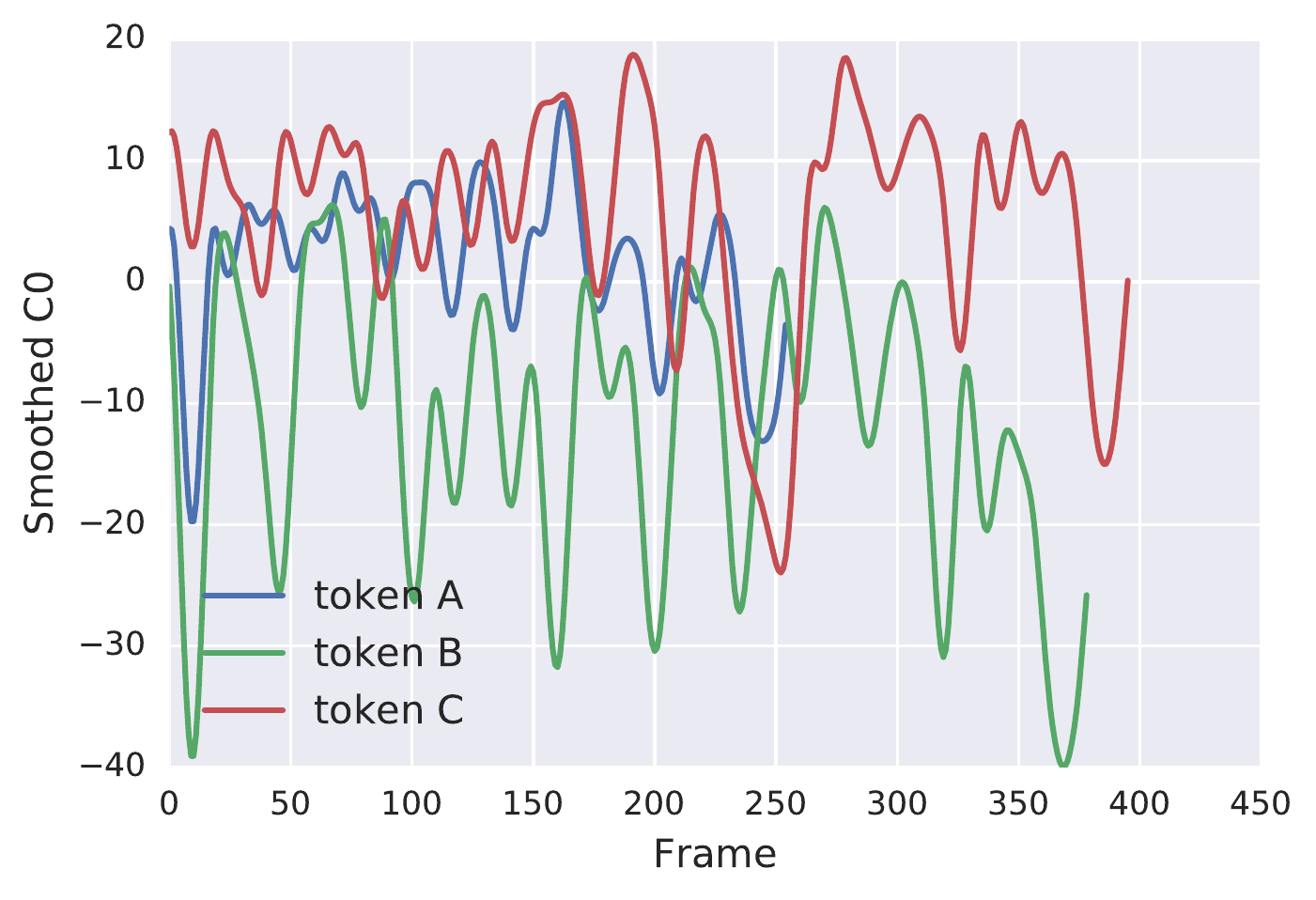}
}
\subfigure[Sentence B]{
\label{fig.tokens.b}
\includegraphics[scale=0.27]{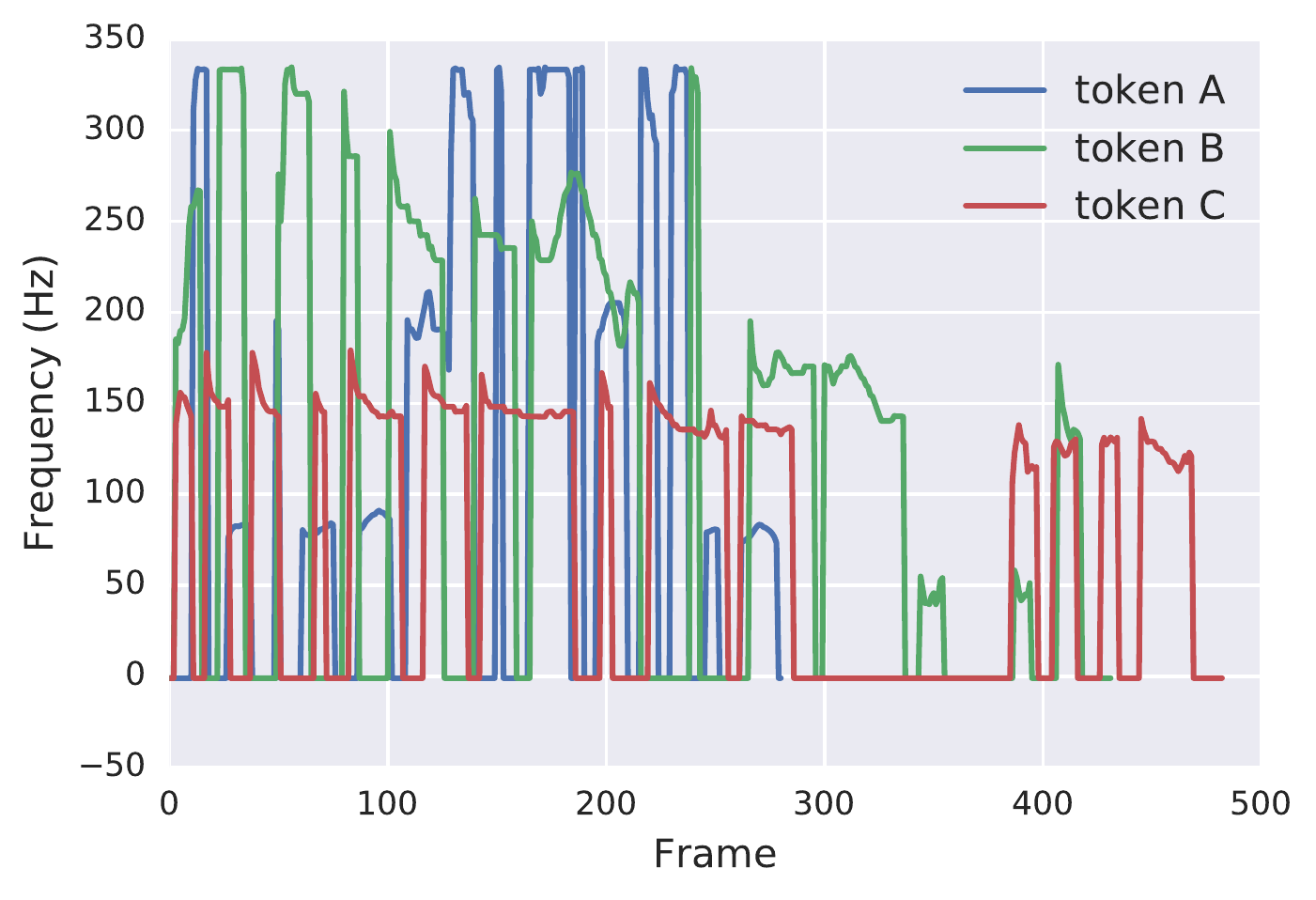}
\includegraphics[scale=0.27]{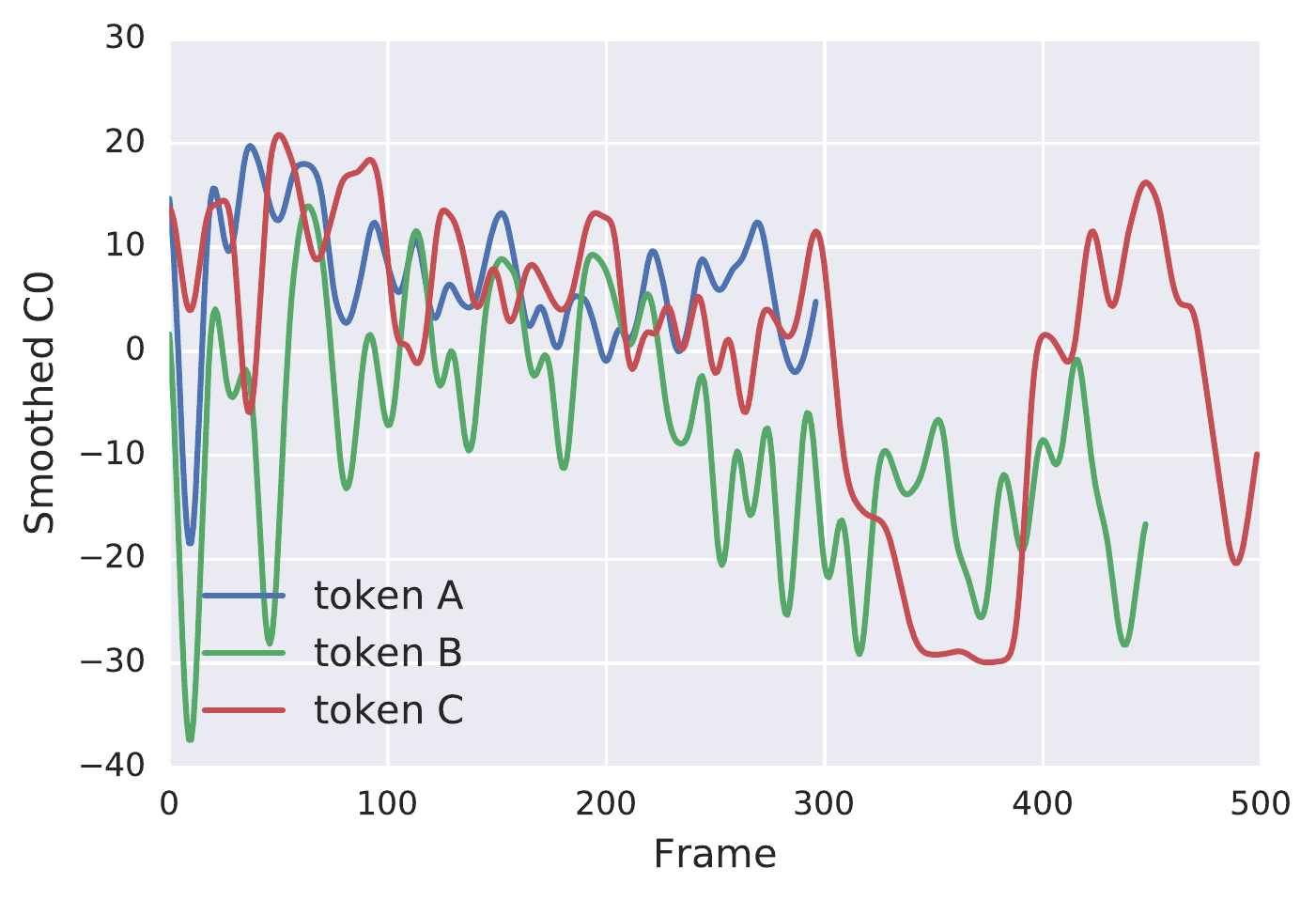}
}
\caption{{\it F0 and C0 (log scale) of two different sentences, synthesized using three tokens. Independent of the text content, the same token exhibits the same F0/C0 trend relative to the other tokens.}}
\vskip -0.2in
\label{fig.tokens.sent_ab}
\end{figure}

\begin{figure*}[t]
\centering
\subfigure[Token A (speed)]{
\label{fig.scaling.speed}
\includegraphics[scale=0.3]{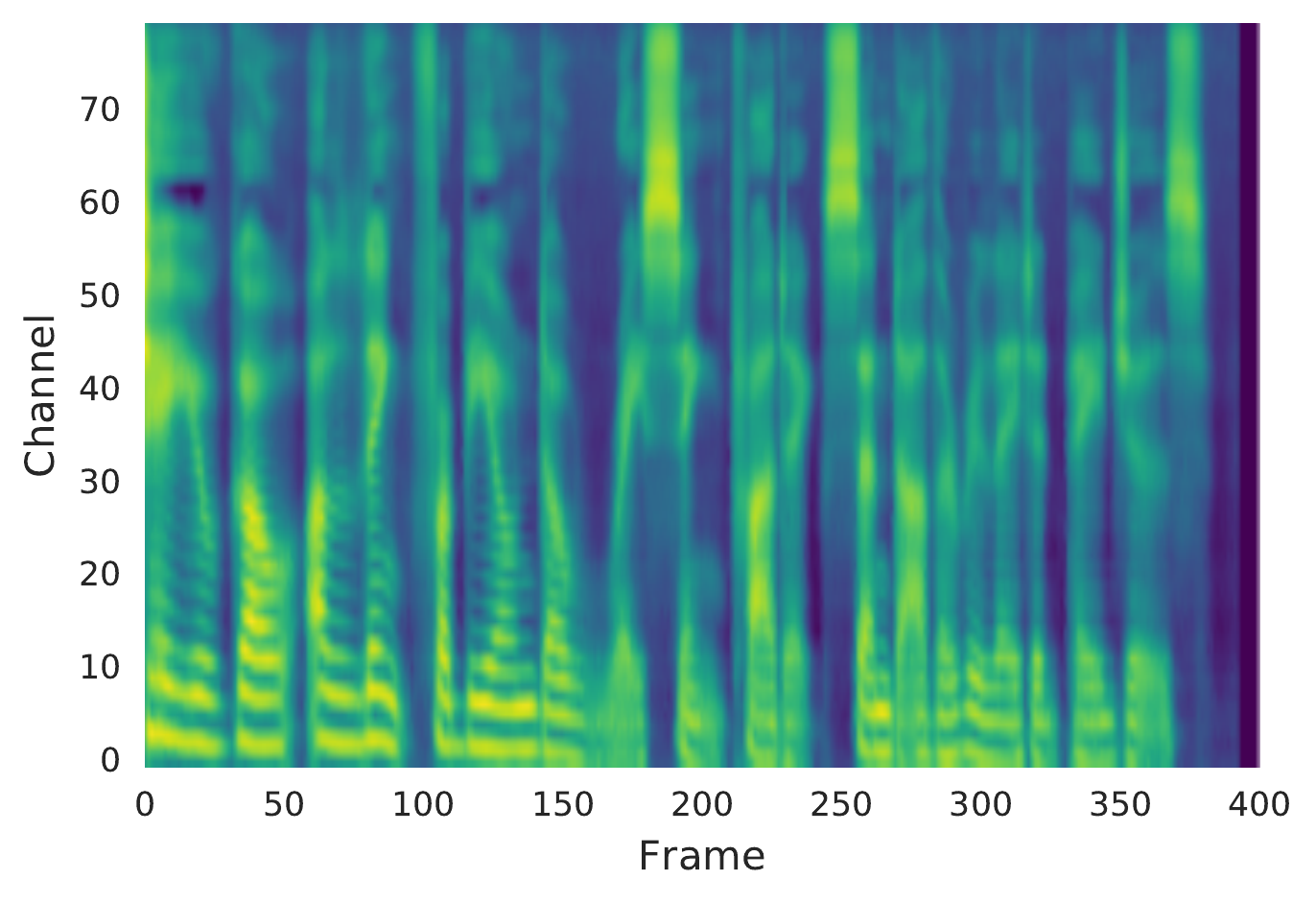}
\includegraphics[scale=0.3]{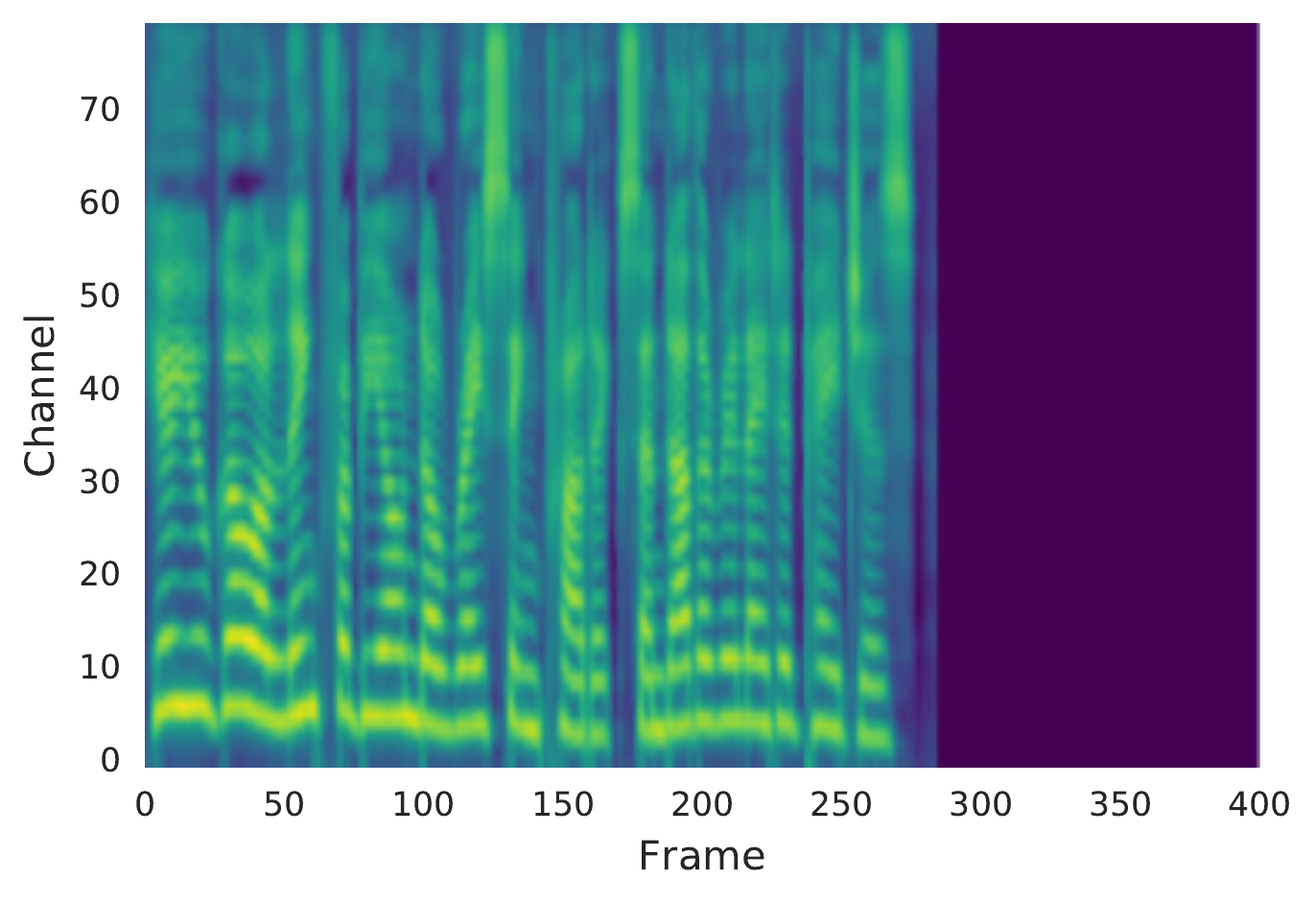}
\includegraphics[scale=0.3]{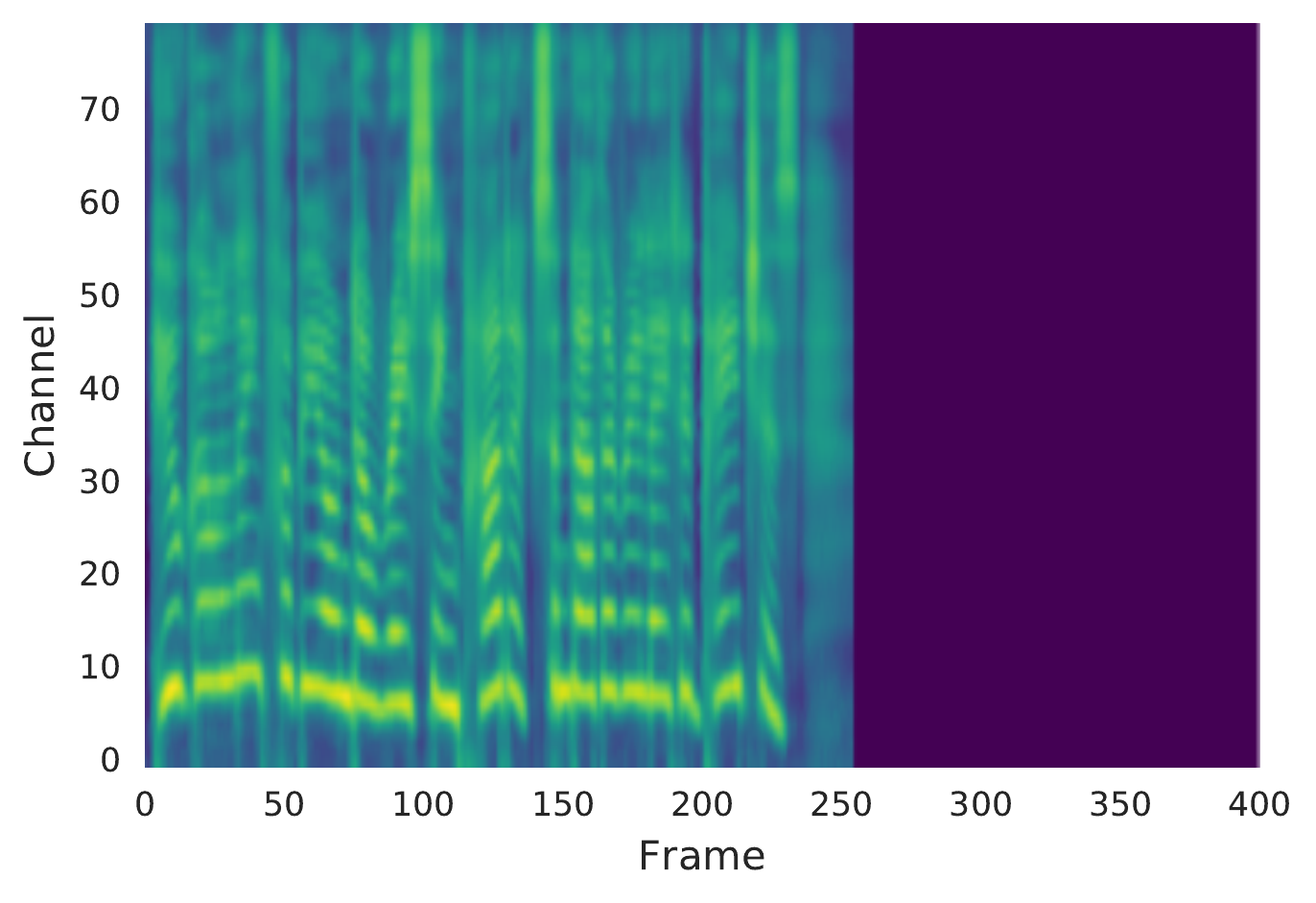}
\includegraphics[scale=0.3]{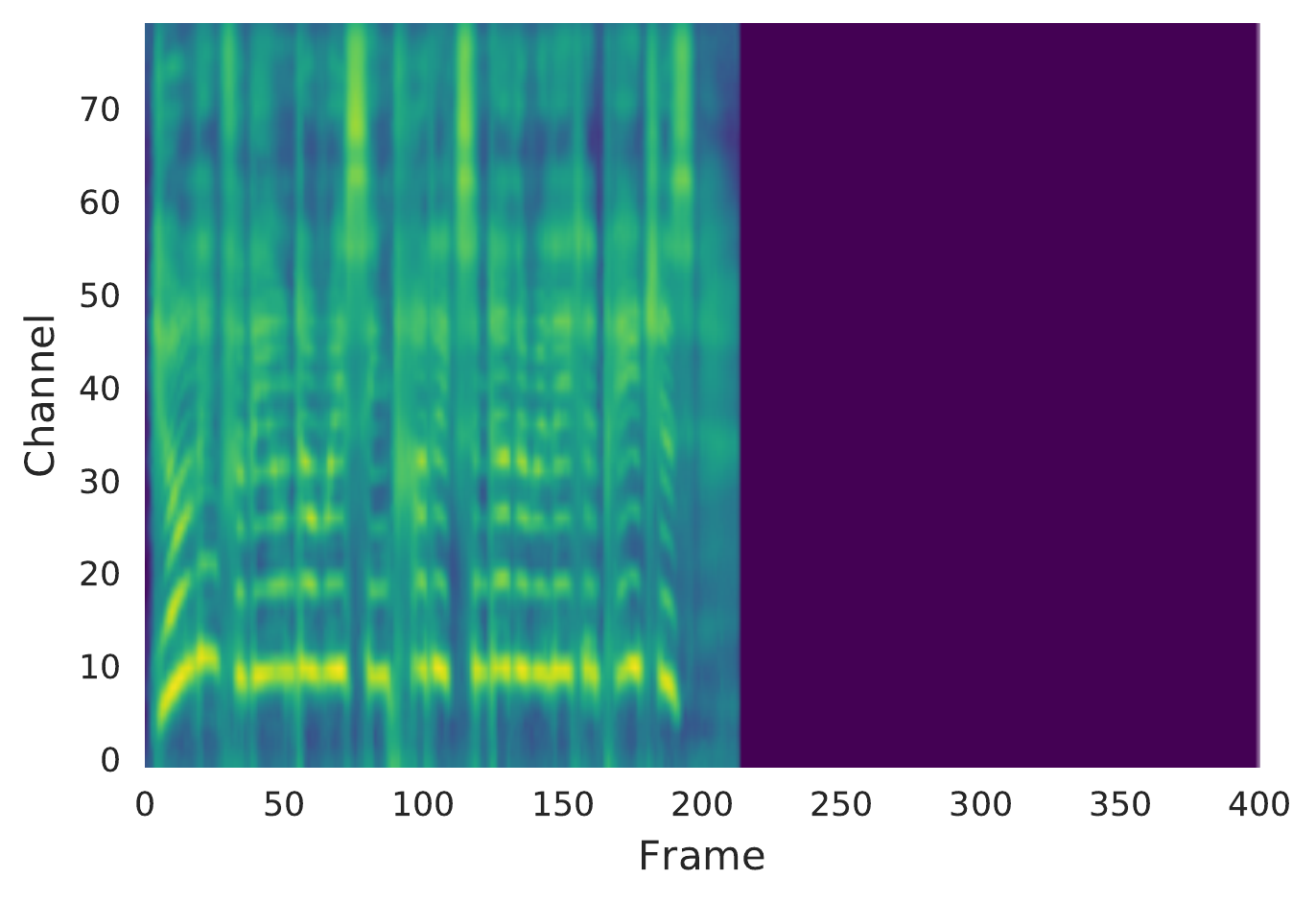}
}
\subfigure[Token B (animated)]{
\label{fig.scaling.exicting}
\includegraphics[scale=0.3]{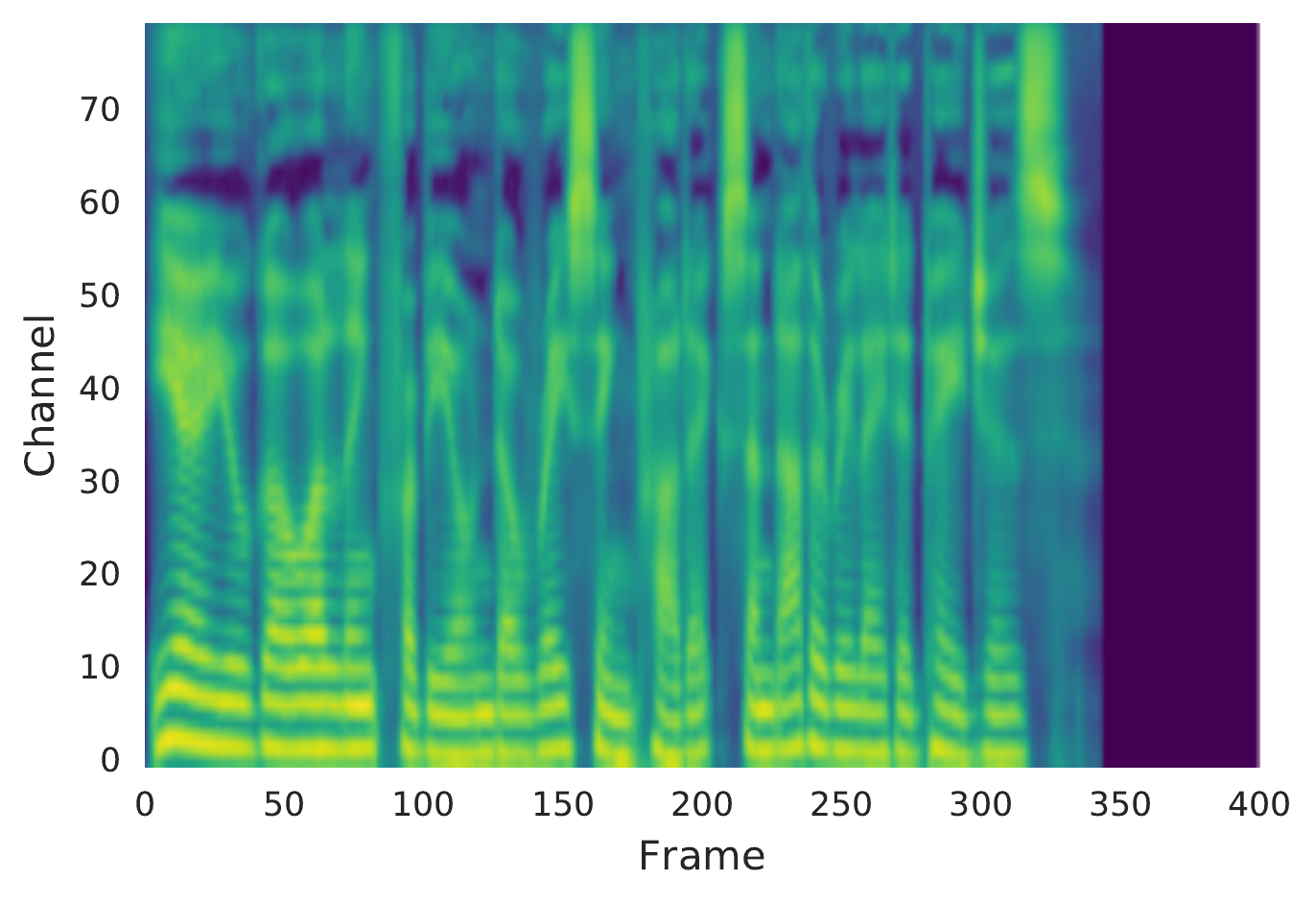}
\includegraphics[scale=0.3]{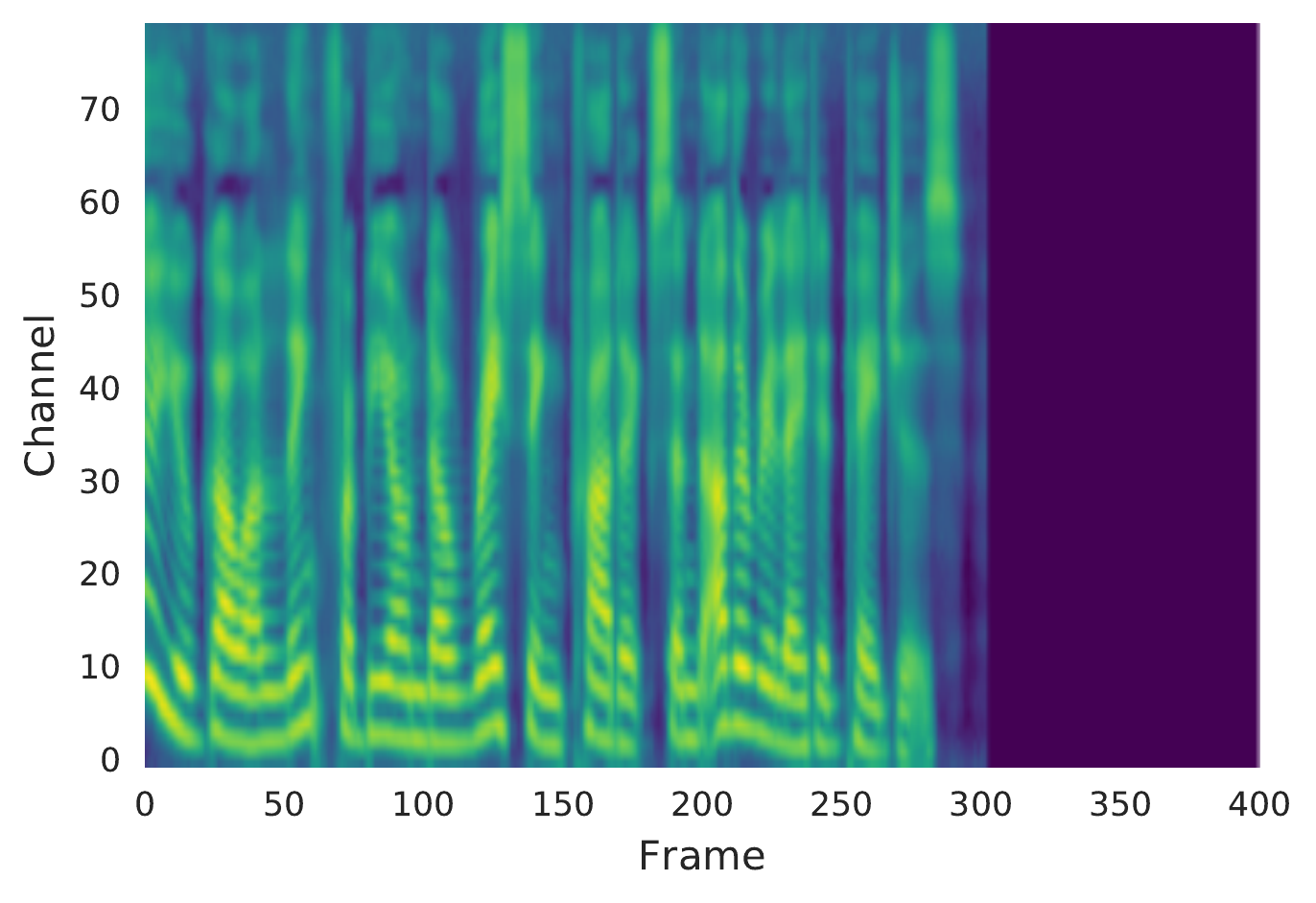}
\includegraphics[scale=0.3]{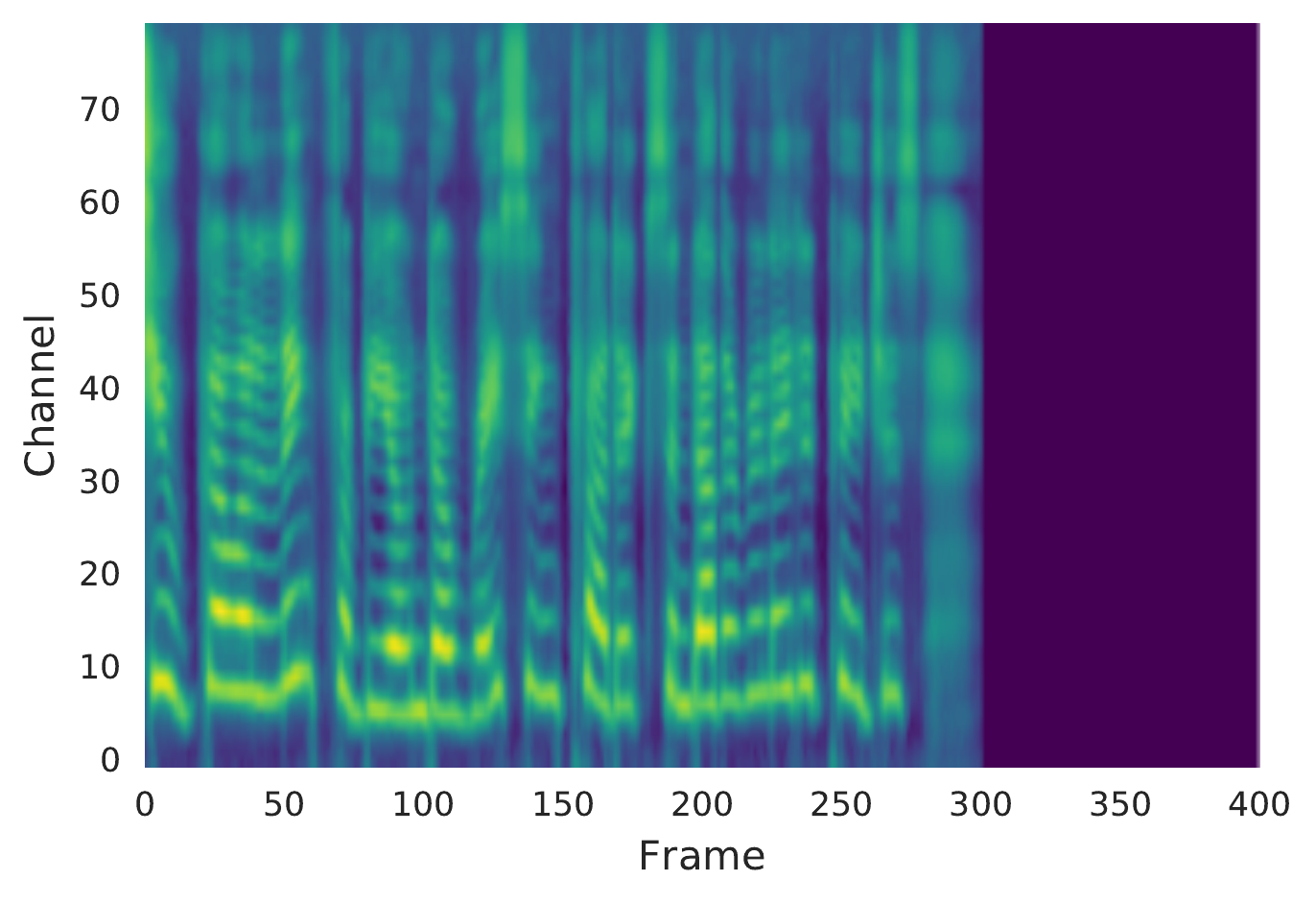}
\includegraphics[scale=0.3]{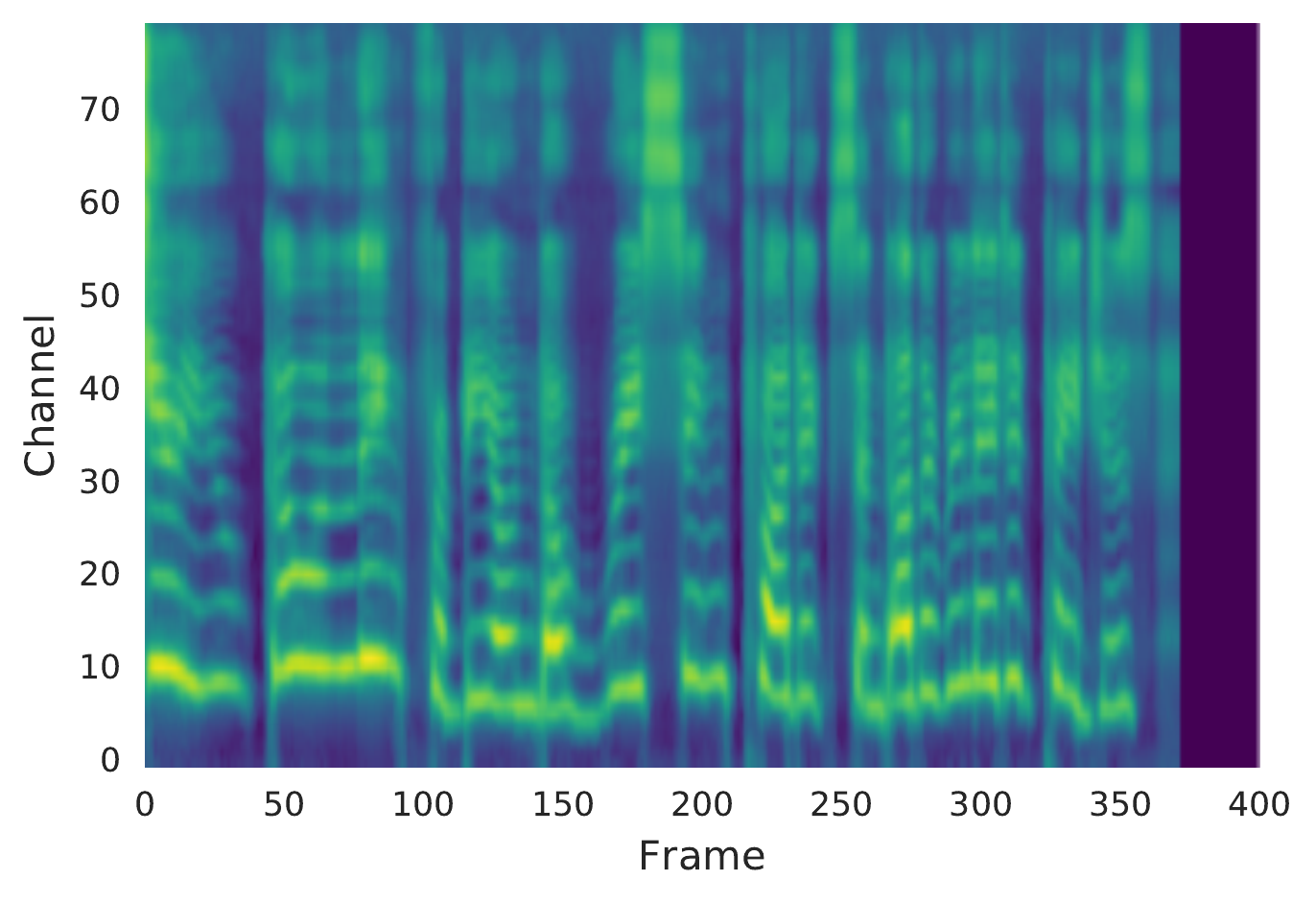}
}
\caption{{\it Effect of token scaling. From left to right, we scale the two tokens by -0.3, 0.1, 0.3, 0.5, respectively. Note that the model seems to exhibit the reverse effect (e.g. fast to slow or animated to calm) with a negative scale, which is never seen during training.}}
\label{fig.scaling_and_speed}
\vskip -0.in
\end{figure*}

\begin{figure}[th!]
\centering
\subfigure[Raw audio] {\includegraphics[scale=0.27]{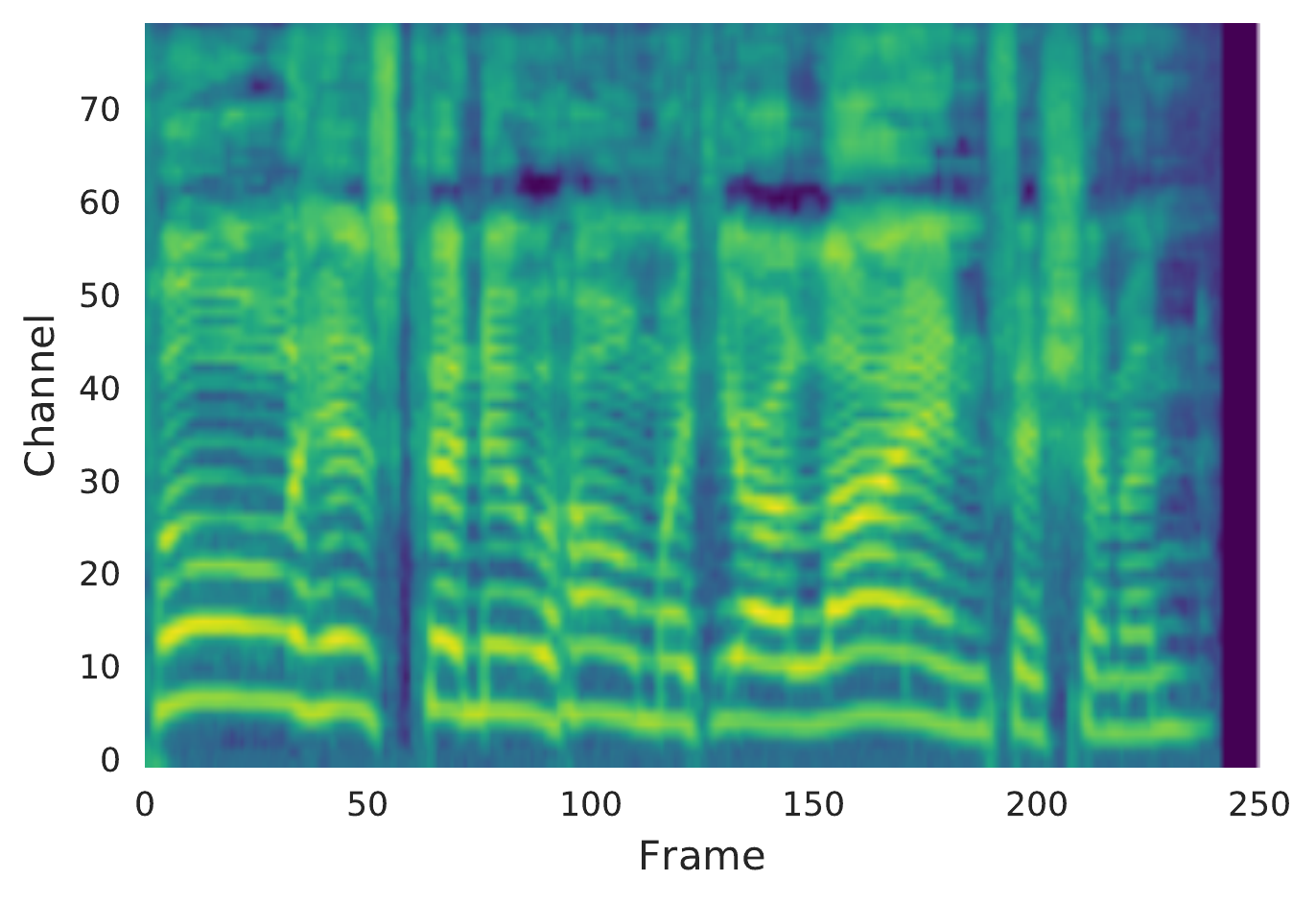}}
\subfigure[Baseline Tacotron]
{\includegraphics[scale=0.27]{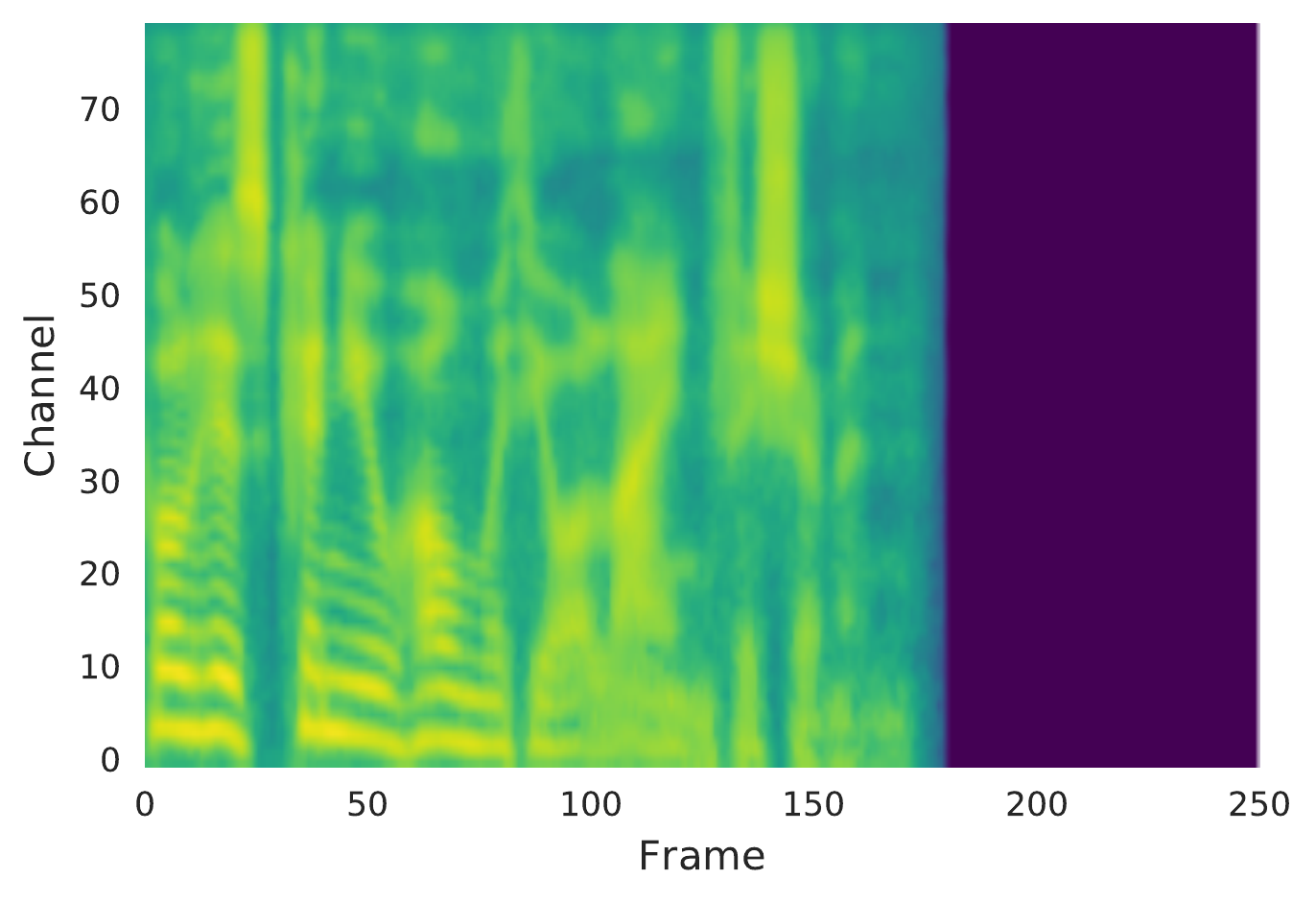}}
\subfigure[Direct conditioning] {\includegraphics[scale=0.27]{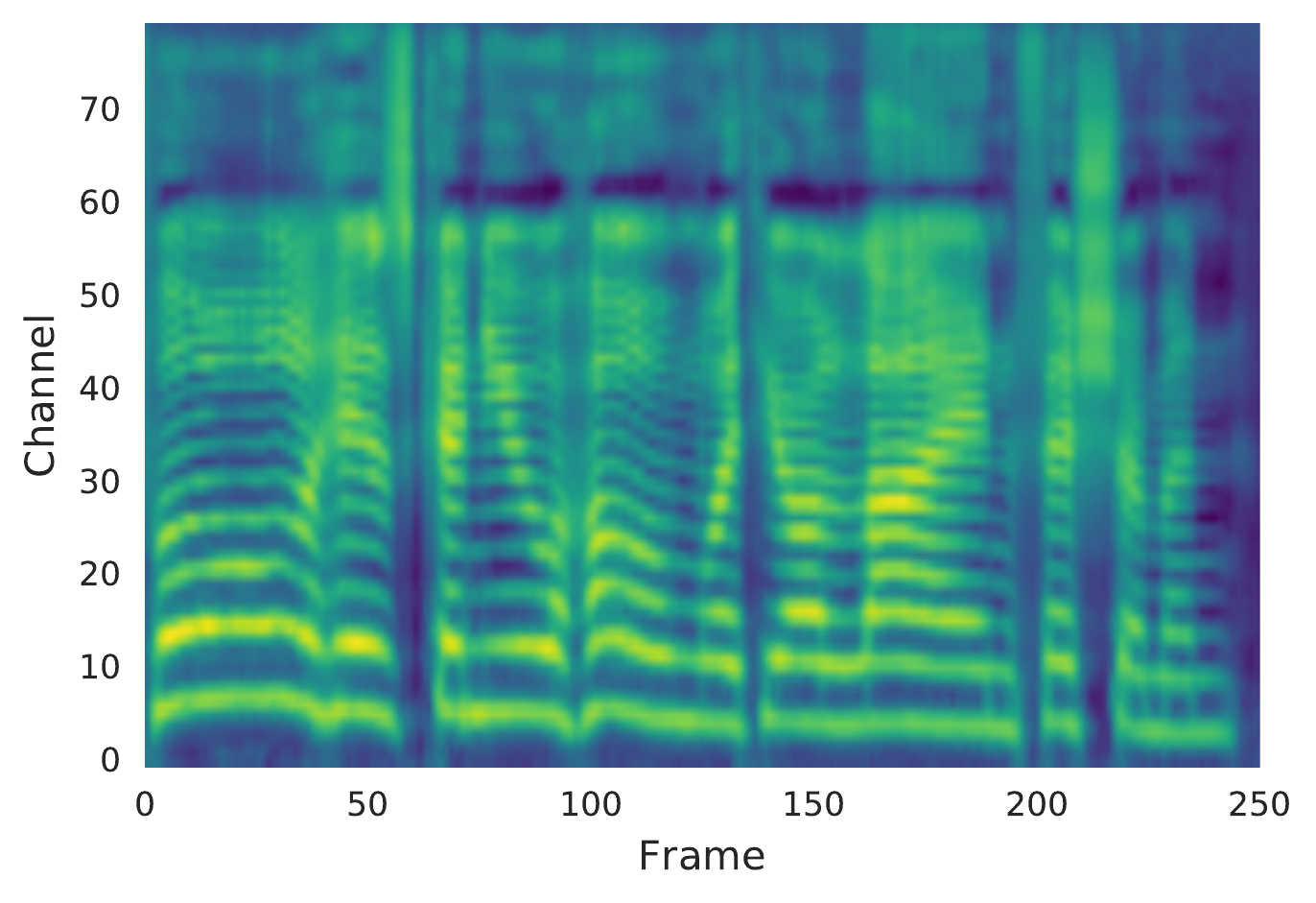}}
\subfigure[GST conditioning]
{\includegraphics[scale=0.27]{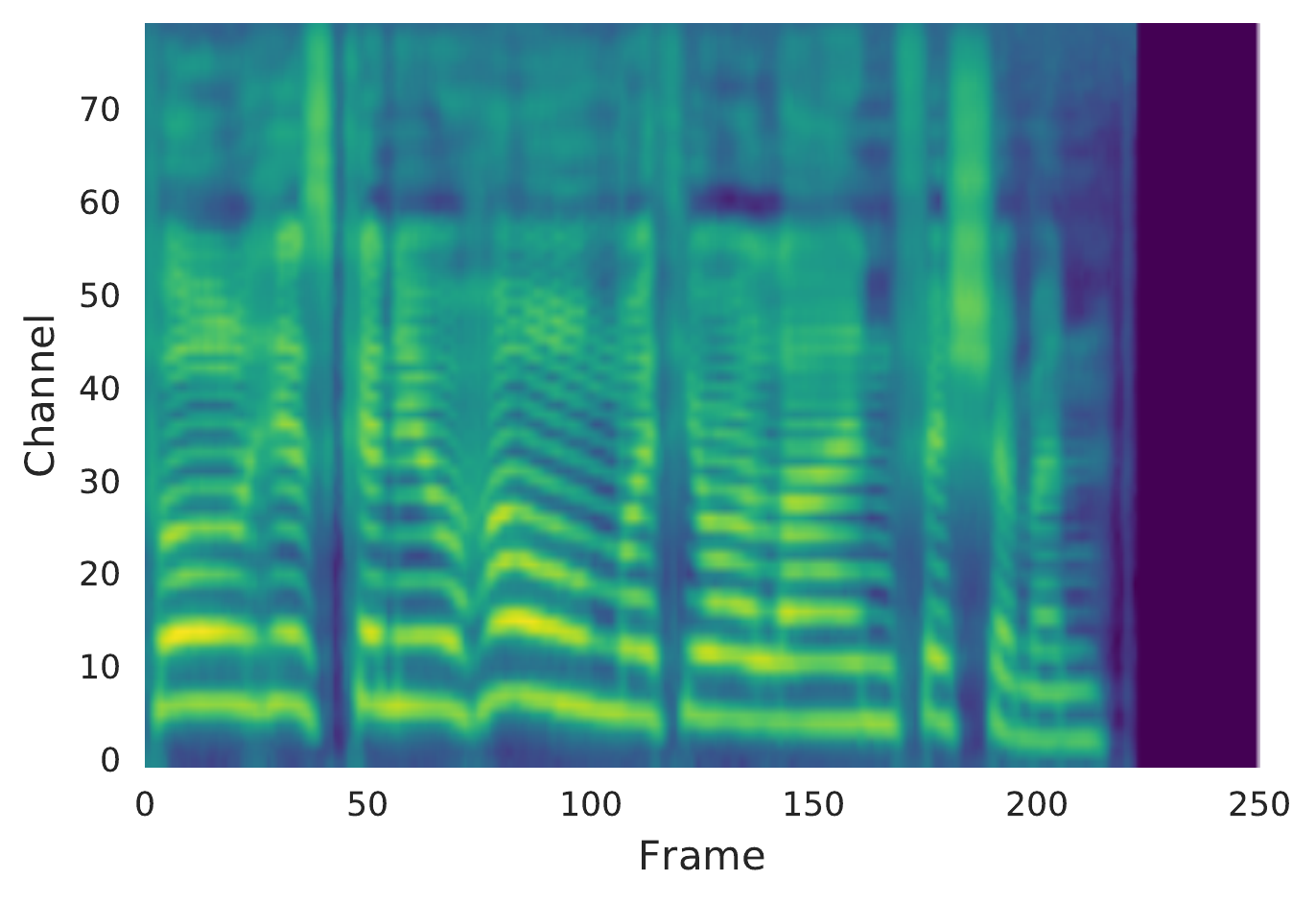}}
\caption{{\it Log-mel spectrograms for parallel style transfer.}}
\label{fig.expts.silt}
\vskip -0.2in
\end{figure}

\subsection{Style Control}
\label{sec.expt.control}

\subsubsection{Style Selection}
\label{sec.expt.control.selection}
The simplest method of control is conditioning the model on an individual token. At inference time, we simply replace the style embedding with a specific, optionally scaled token. 

Conditioning in this manner has several benefits. First, it allows us to examine which style attributes each token encodes. Empirically, we find that each token can represent not just pitch and intensity, but also a variety of other attributes, such as speaking rate and emotion. This can be seen in Figure \ref{fig.tokens.sent_ab}, which shows two sentences synthesized with three different style tokens (scale=0.3) from a 10-token GST model. The plots show that F0 and C0 (energy) curves are quite different across style tokens. However, the F0 and C0 contours generated by each token follow a clear relative trend, despite the fact that input sentences A and B are completely different.

Indeed, perceptually, the red token corresponds to a lower-pitch voice, the green token to a decreasing pitch, and the blue token to a faster speaking rate (note the total audio duration in both plots).

Single-token conditioning also reveals that not all tokens capture single attributes: while one token may learn to represent speaking rate, others may learn a mixture of attributes that reflect stylistic co-occurrence in the training data (a low-pitched token, for example, can also encode a slower speaking rate). Encouraging more independent style attribute learning is an important focus of ongoing work.

In addition to providing interpretability, style token conditioning can also improve synthesis quality. Consider the problem of long-form synthesis on training data with lots of prosodic variation. Many TTS models learn to generate the ``average'' prosodic style, which can be problematic for expressive datasets, since the very variation that characterizes them is collapsed. This can also lead to undesirable side effects, such as pitch continuously declining towards the end of each sentence. We find that conditioning on ``lively''-sounding tokens can address both of these problems, significantly improving the prosodic variation.

Audio examples of style selection can be found \href{https://google.github.io/tacotron/publications/global_style_tokens/index.html#style_control.selection}{here}.

\subsubsection{Style scaling}
Another method for controlling style token output is via scaling.
We find that multiplying a token embedding by a scalar value intensifies
its style effect. (Note that large scaling values may lead to unintelligible speech, which suggests future work on improving stability.) This is illustrated in Figure \ref{fig.scaling_and_speed}, which shows spectrograms of utterances synthesized by two different tokens. Perceptually, these tokens encode two different speaking styles: a faster speaking rate (\ref{fig.scaling.speed}), and more animated speech (\ref{fig.scaling.exicting}). Figure \ref{fig.scaling.speed} shows that increasing the scaling factor of the faster speaking rate token causes a gradual compression of the spectrogram in the time domain. Similarly, Figure \ref{fig.scaling.exicting} shows that increasing the scaling factor of the animated speech token yields commensurate increases in pitch variation. 
These style scaling effects hold even for negative values (speaking rate becomes slower, and speech becomes calmer), despite the fact that the model only sees positive (softmax) values during training.

Audio examples of style scaling can be found \href{https://google.github.io/tacotron/publications/global_style_tokens/index.html#style_control.scaling}{here}.

\subsubsection{Style Sampling}

We can also control synthesis during inference by modifying the attention module weights inside the style token layer. Since the GST attention produces a set of combination weights, these may be refined manually to yield a desired interpolation. We can also use randomly generated softmax weights to sample the style space. The sampling diversity can be controlled by tuning the softmax temperature.

\subsubsection{Text-side style control/morphing}
While the same style embedding is added to all text encoder states during training, this doesn't need to be the case in inference mode. As our audio samples demonstrate, this allows us to do piecewise style control or morphing by conditioning on one or more tokens for different segments of input text. 

Audio examples of style morphing can be found \href{https://google.github.io/tacotron/publications/global_style_tokens/index.html#style_control.morphing}{here}.

\subsection{Style Transfer}
\label{sec.expt.transfer}

Style transfer is an active area of research that aims to synthesize a phrase in the prosodic style of a reference signal \cite{Zhizheng13,Nakashika2016,Tomi2017VC}. The property that a GST model can be conditioned on any convex combination of style tokens lends itself well to this task;
at inference time (method 2 of Section \ref{sec.model.inference}), we can simply feed
a reference signal to guide the choice of token combination weights.
The experiments below use 4-head GST attention.

\begin{figure*}[t]
\centering
\subfigure[10-token GST]{
\label{fig.robustness.gst}
\includegraphics[scale=0.29]{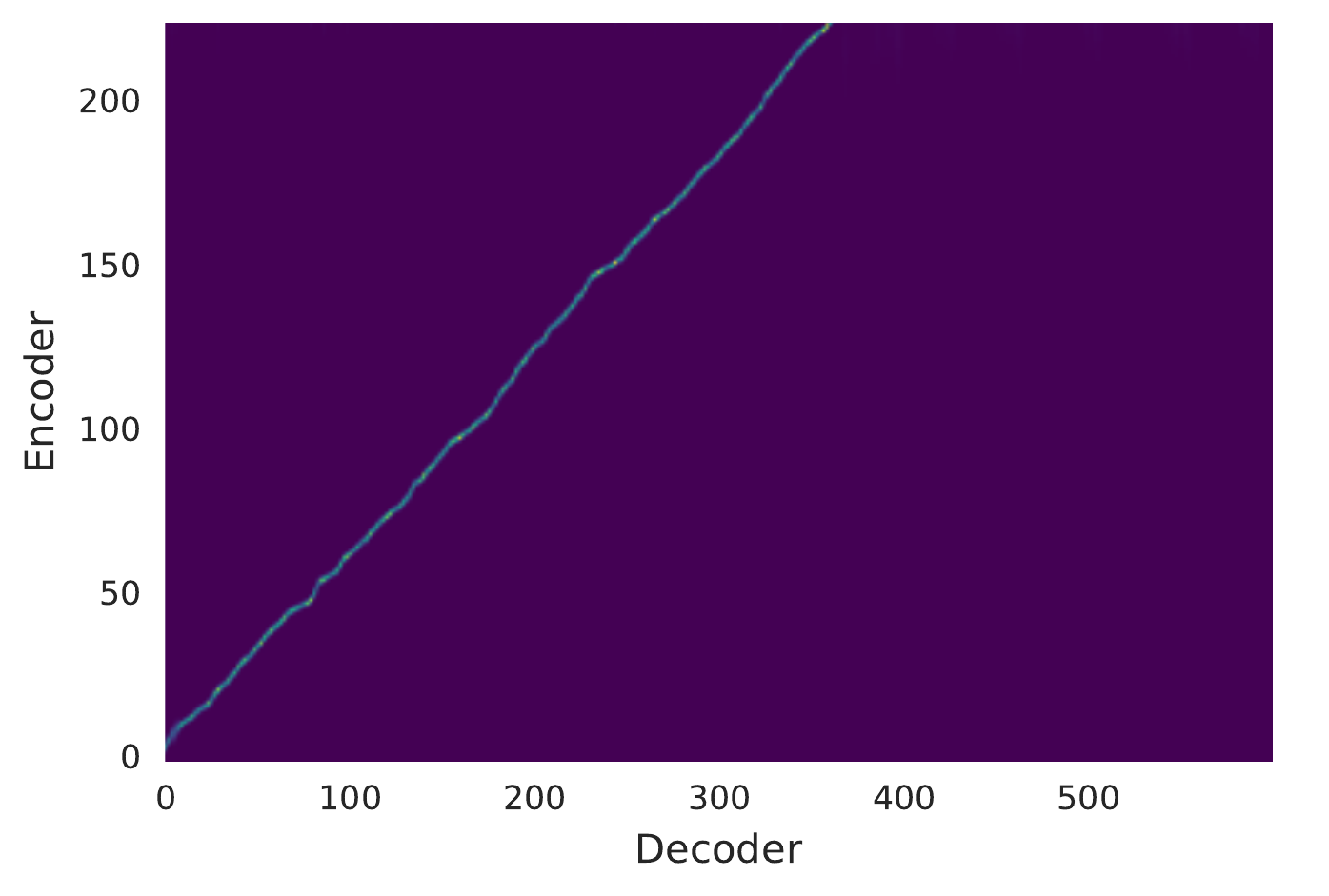}
\includegraphics[scale=0.29]{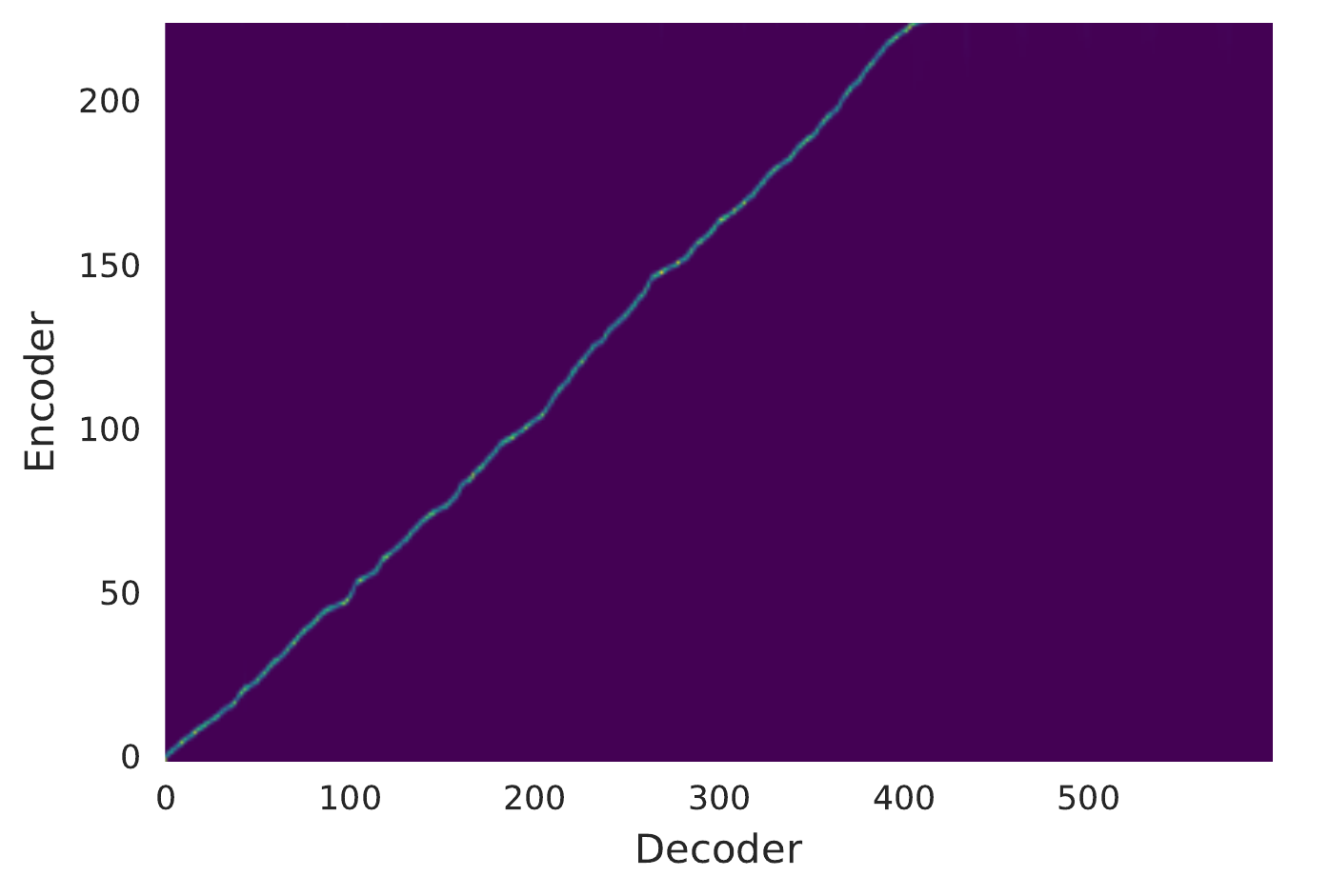}
\includegraphics[scale=0.29]{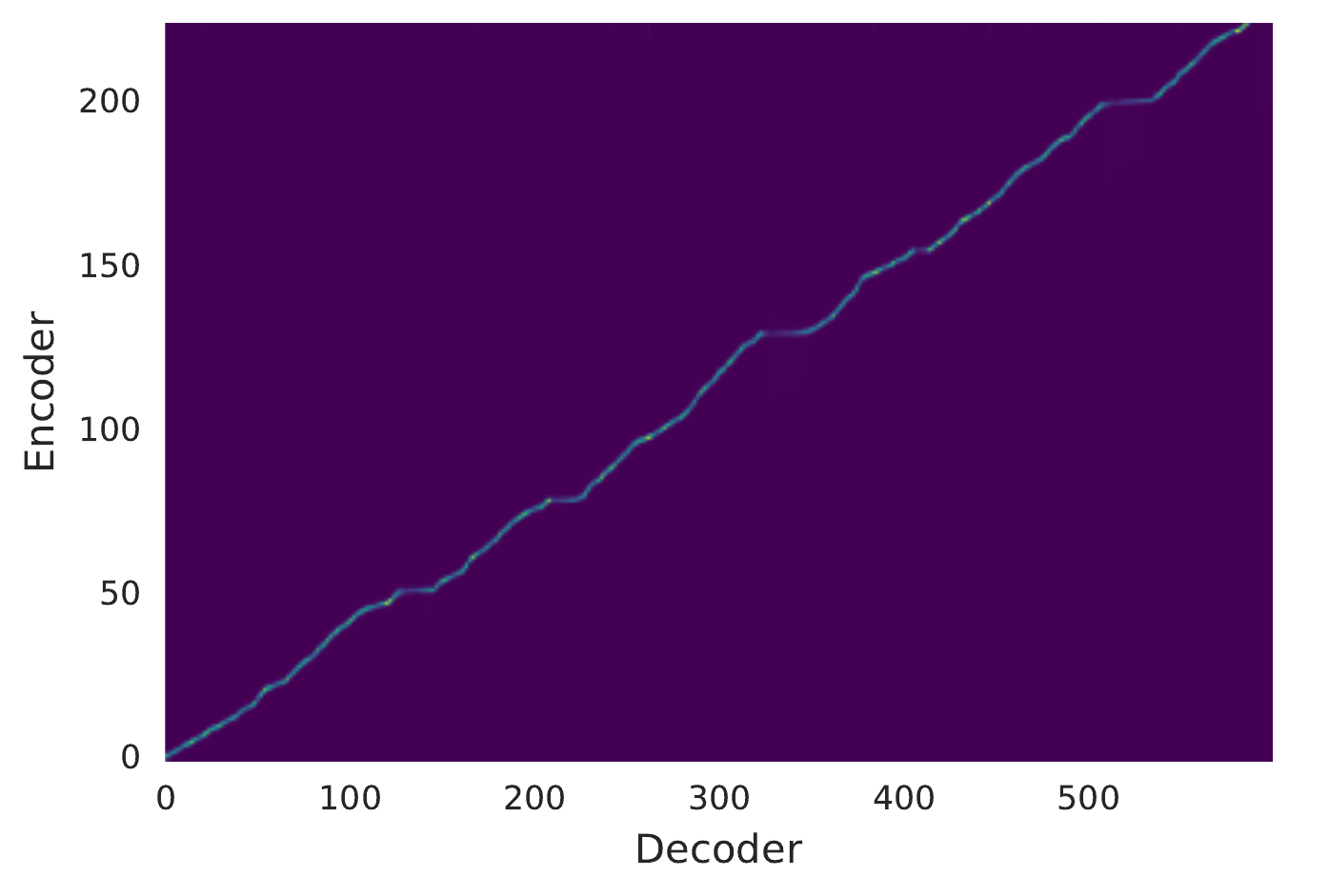}
}
\vskip -0.05in
\subfigure[Direct conditioning (128-D)]{
\label{fig.robustness.nogst}
\includegraphics[scale=0.29]{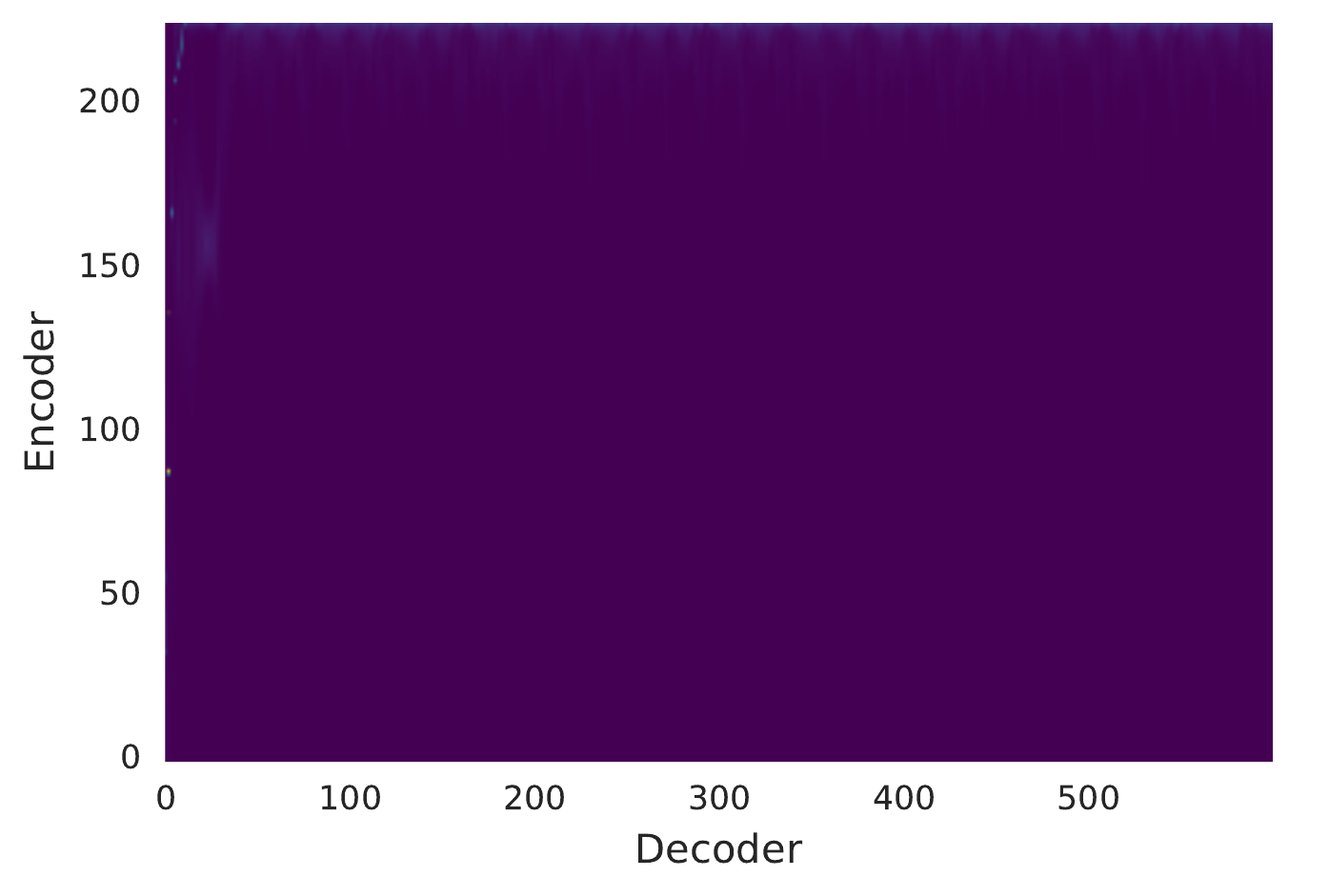}
\includegraphics[scale=0.29]{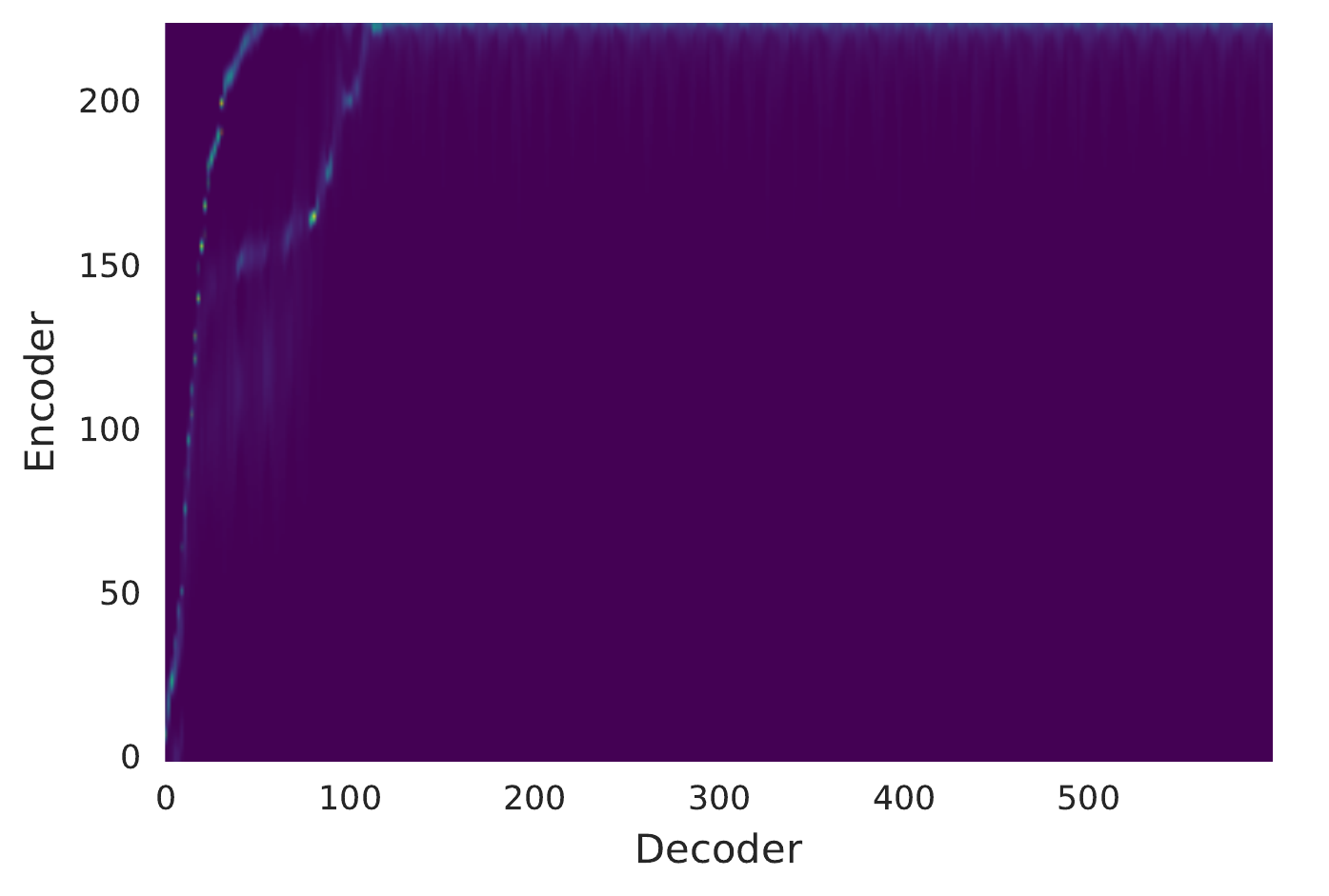}
\includegraphics[scale=0.29]{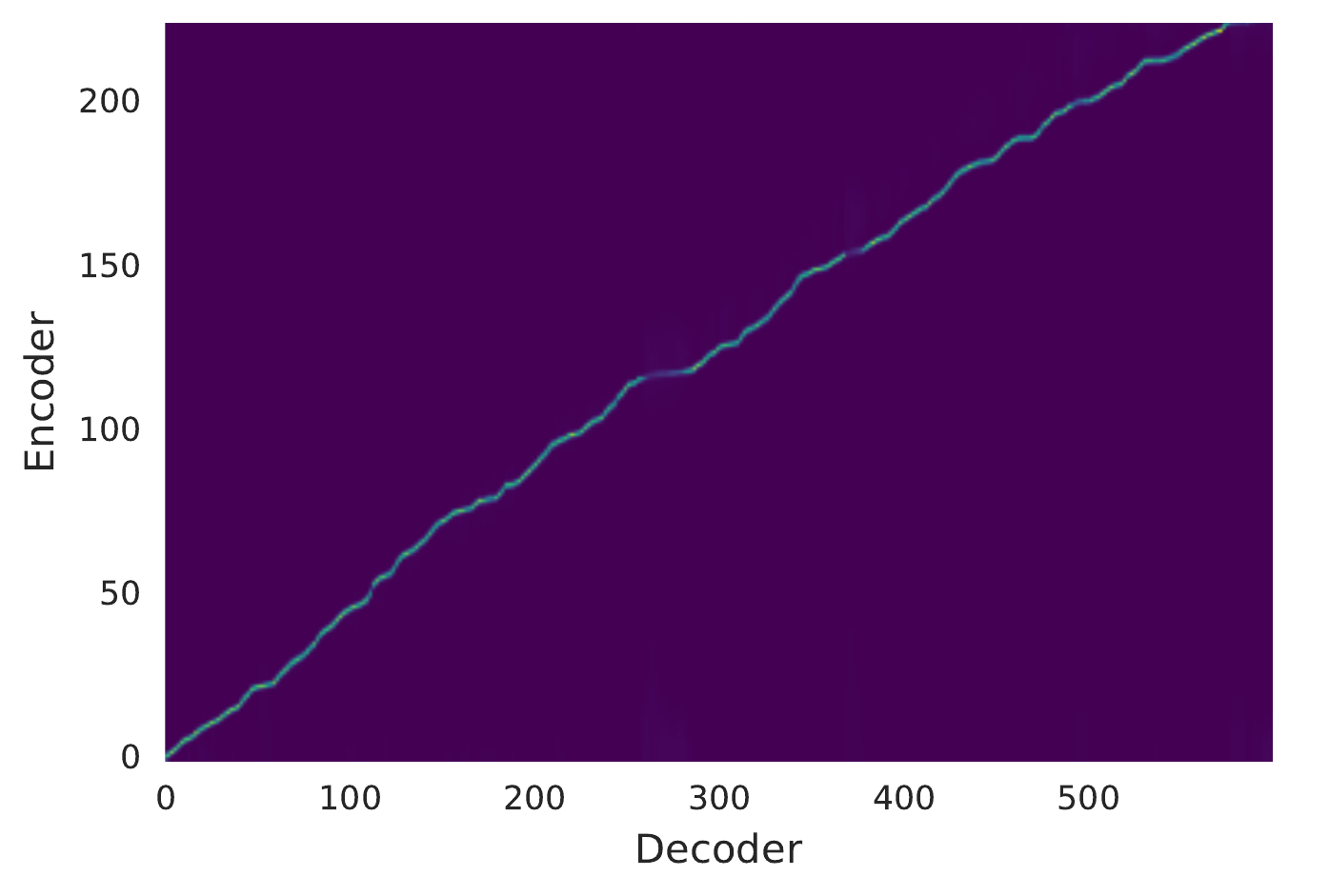}
}
\vskip -0.05in
\subfigure[256-token GST]{
\label{fig.robustness.gst256}
\includegraphics[scale=0.29]{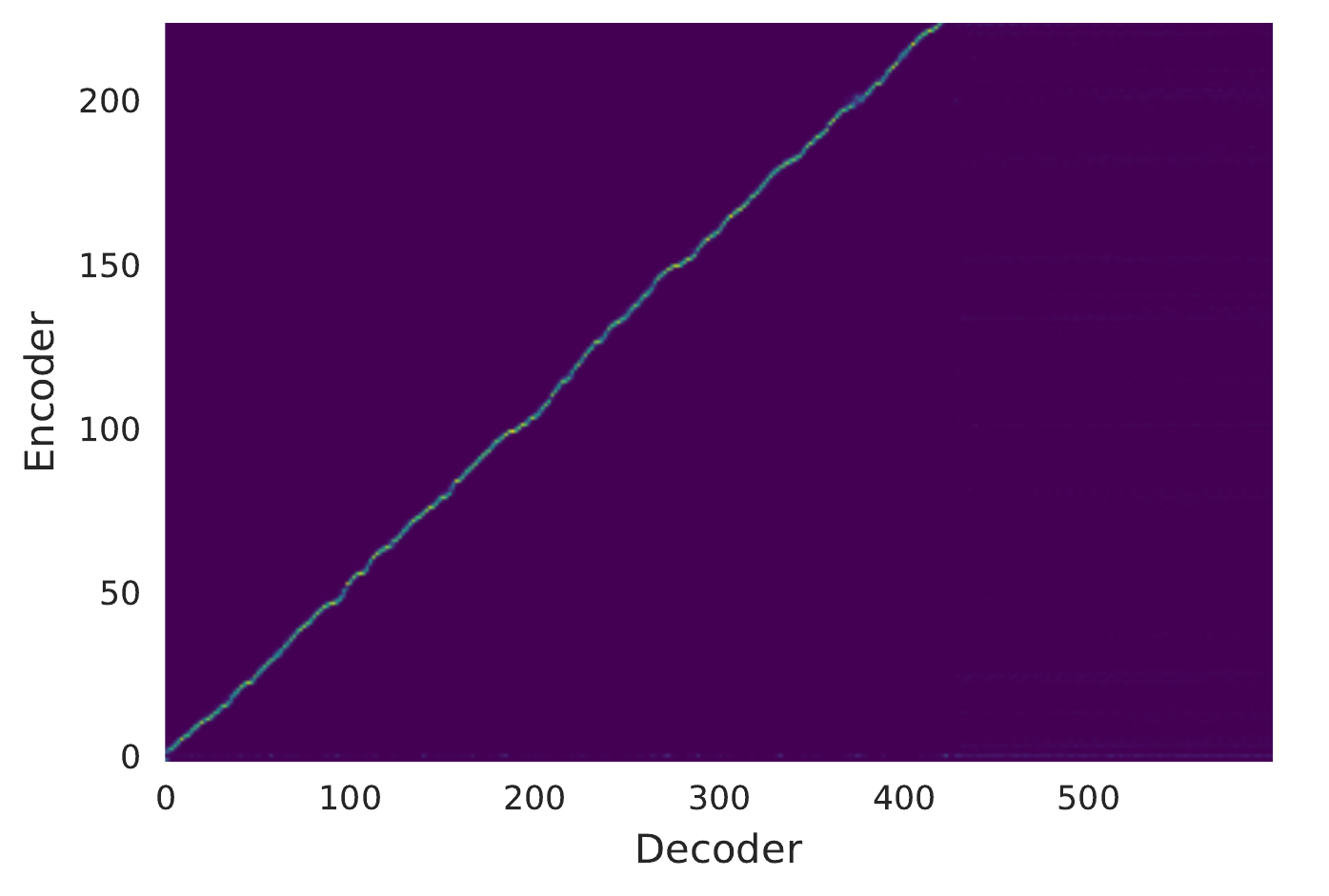}
\includegraphics[scale=0.29]{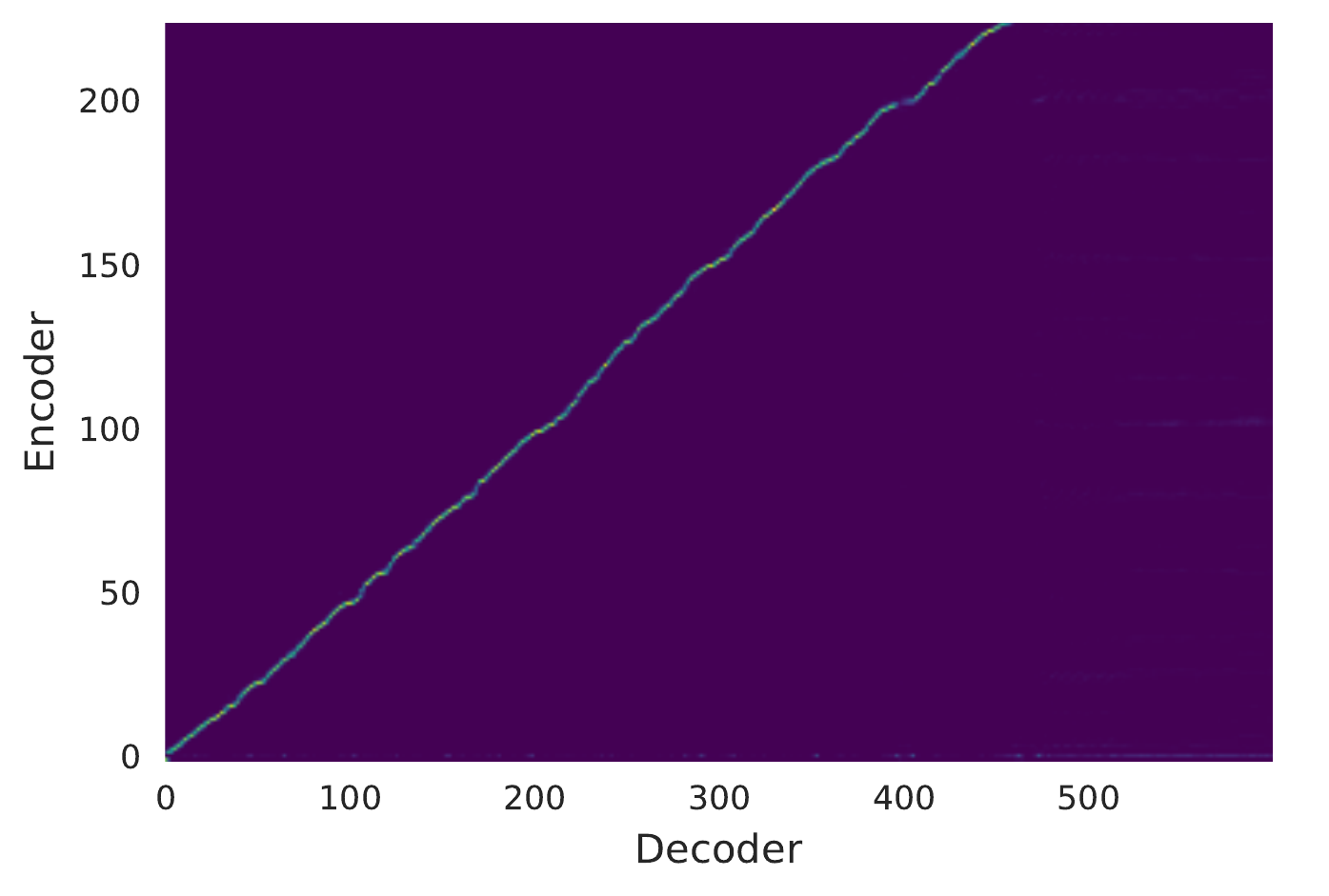}
\includegraphics[scale=0.29]{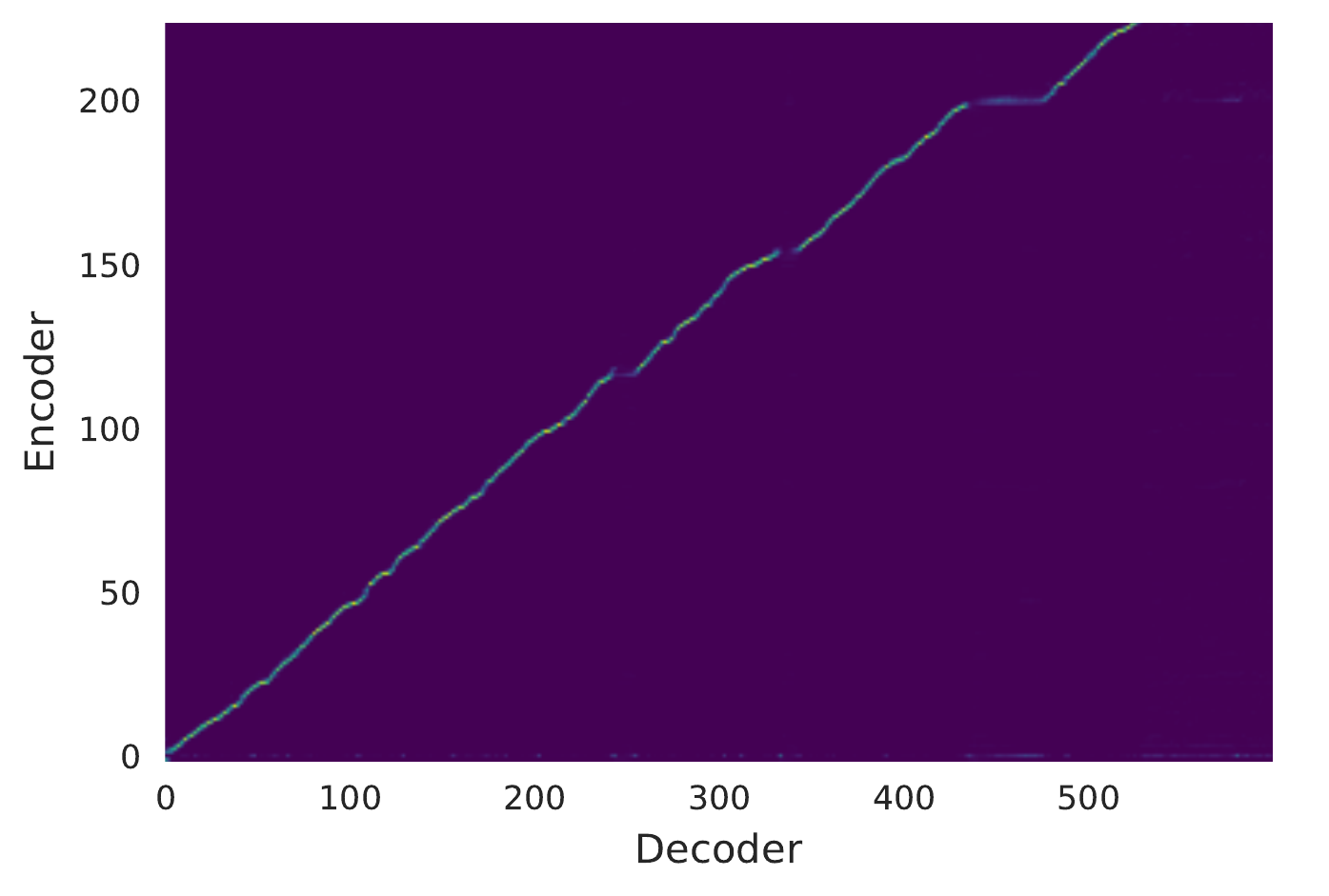}
}
\caption{{\it Robustness in non-parallel style transfer. Left to right: attention alignments obtained from feeding three references whose text lengths are 10, 96, 321 characters, respectively. The target text length is 258 characters.}}
\label{fig.expts.non_parallel_transfer}
\vskip -0.1in
\end{figure*}

\subsubsection{Parallel Style Transfer}
\label{sec.expt.transfer.parallel}
Figure \ref{fig.expts.silt} shows spectrograms for a \textit{parallel transfer} task, where the text to synthesize matches the text of the reference signal. The GST model spectrogram is at the bottom right, compared to three other baselines: (a) the ground-truth input signal (i.e. the reference); (b) inference performed by a baseline Tacotron model (which infers acoustics only from text); and (c) inference as performed by \cite{rj2018transfer}, a Tacotron system which conditions the text encoder directly on an 128-D reference embedding.

We see that, given only text input, the baseline Tacotron model does not closely match the prosodic style of the reference signal. By contrast, the direct conditioning method of \cite{rj2018transfer} results in nearly time-aligned fine prosody transfer. The GST model is somewhere in between: while its output duration and formant transitions don't precisely match those of the reference, the overall spectrotemporal envelopes do. Perceptually, GSTs resemble the prosodic style of the reference.

Audio examples of parallel style transfer can be found \href{https://google.github.io/tacotron/publications/global_style_tokens/index.html#style_transfer.parallel}{here}.

\subsubsection{Non-Parallel Style Transfer}
\label{sec.expt.transfer.nonparallel}
We next show results for a \textit{non-parallel transfer} task, in which a TTS system must synthesize arbitrary text in the prosodic style of a reference signal. We chose three different reference signals for this task, and tested how well a GST model replicated each style when synthesizing the same target phrase.
Since long-form synthesis can benefit significantly from proper stylistic rendering, we used a long (258-character) target phrase. We chose source phrases of varying lengths (10, 96, and 321 characters, respectively).
Figure \ref{fig.expts.non_parallel_transfer} shows alignment matrices 
for synthesis conditioned on each source signal.

The top row shows a 10-token GST model. This model robustly generalizes to all three conditioning inputs, as evidenced by the good alignment plots. The bottom row shows a 256-token GST model exhibiting the same behavior; we include this model to show that  GSTs remain robust even when the number of tokens (256) is larger than the reference embedding dimensionality (128).

The middle row shows a model with direct reference embedding conditioning. The attention matrices show that this model fails when conditioned on the shorter source phrases, since it tries to squeeze its synthesis into the same time interval as that of the reference. While the model successfully aligns when conditioned on the longest input, intelligibility is poor for some words: the per-utterance embedding captures too much information (such as timing and phonetics) from the source, hurting generalization.

To evaluate the quality of this method at scale, we ran side-by-side subjective tests of non-parallel GST style transfer against a Tacotron baseline. We used an evaluation set of 60 audiobook sentences, including many long phrases. We generated two sets of GST output by conditioning the model on two different narrative-style reference signals, unseen during training. A side-by-side subjective test indicated that raters preferred both sets of GST synthesis against a Tacotron baseline, as shown in Table \ref{tb.sxs}. 

The performance of GSTs on non-parallel style transfer is significant, since it allows using a source signal to guide robust stylistic synthesis of arbitrary text.

Audio examples of non-parallel style transfer can be found \href{https://google.github.io/tacotron/publications/global_style_tokens/index.html#style_transfer.nonparallel}{here}.

\begin{table}[t!]

\caption{SxS subjective
preference (\%) and $p$-values
of GST audiobook synthesis against a Tacotron baseline. Each row shows GST inference conditioned a different reference signal (A and B). $p$-values are given for both a 3-point and 7-point rating system.}
\label{tb.sxs}
\begin{center}
\begin{small}
\begin{sc}
\setlength\tabcolsep{3.5pt} 
\begin{tabular}{c|ccc|cc}
\toprule
& \multicolumn{3}{c|}{preference (\%)} & \multicolumn{2}{c} {p-value}
\\ \cline{2-6}
& Base & neutral & GST & 3-point & 7-point \\
\midrule
Signal A & 32.9 & 26.5 & 40.6 & p=0.0552 & p=0.0131 \\
Signal B & 33.1 & 21.9 & 45.0 & p=0.0038 & p=0.0003 \\
\bottomrule
\end{tabular}
\end{sc}
\end{small}
\end{center}
\vskip -0.1in
\end{table}

%% file: 7-experiments-noisy.tex
\section{Experiments: Unlabeled Noisy Found Data}
\label{sec.expt.found}
Studio-quality data can be both economically and time consuming to record. While the internet holds vast amounts of rich real-life expressive speech, it is often noisy and difficult to label. 
In this section, we demonstrate how GSTs can be used to train robust models directly from noisy found data, without modifications.

\subsection{Artificial Noisy Data}

\begin{table}[t]
\caption{Robust MOS as a function of the percentage of interference in the training set. The total training set size is the same.}
\label{tb.mtr.mos}
\vskip 0.15in
\begin{center}
\begin{sc}
\begin{tabular}{ccc}
\toprule
Noise \% & Baseline Tacotron & GST \\
\midrule
50\%    & 2.819 $\pm$ 0.269 \ & 4.080 $\pm$ 0.075 \\
75\% & 1.819 $\pm$ 0.227 & 3.993 $\pm$ 0.074 \\
90\% & 1.609 $\pm$ 0.131 & 4.031 $\pm$ 0.082 \\
95\% & 1.353 $\pm$ 0.090 & 3.997 $\pm$ 0.066 \\
\bottomrule
\end{tabular}
\end{sc}
\end{center}
\vskip -0.1in
\end{table}

As a first experiment, we artificially generate training sets by adding noise to clean speech. 
The motivation here is to simulate real noisy data while performing controlled experiments. 
To achieve this, we pass the single-speaker US English proprietary dataset from 
\cite{yx2017tacotron} 
into a room simulator \cite{kim2017generation}, which adds varying types of background noise and room reverberations. The signal-to-noise ratio (SNR) ranges from 5-25 dB, and the T60s of room reverberation ranges from 100-900 ms. We create four different training sets where 50\%, 75\%, 90\% and 95\% of the input is noisified, respectively.

\begin{figure}[t!]
\centering
\subfigure[``Music'' token] {\includegraphics[scale=0.28]{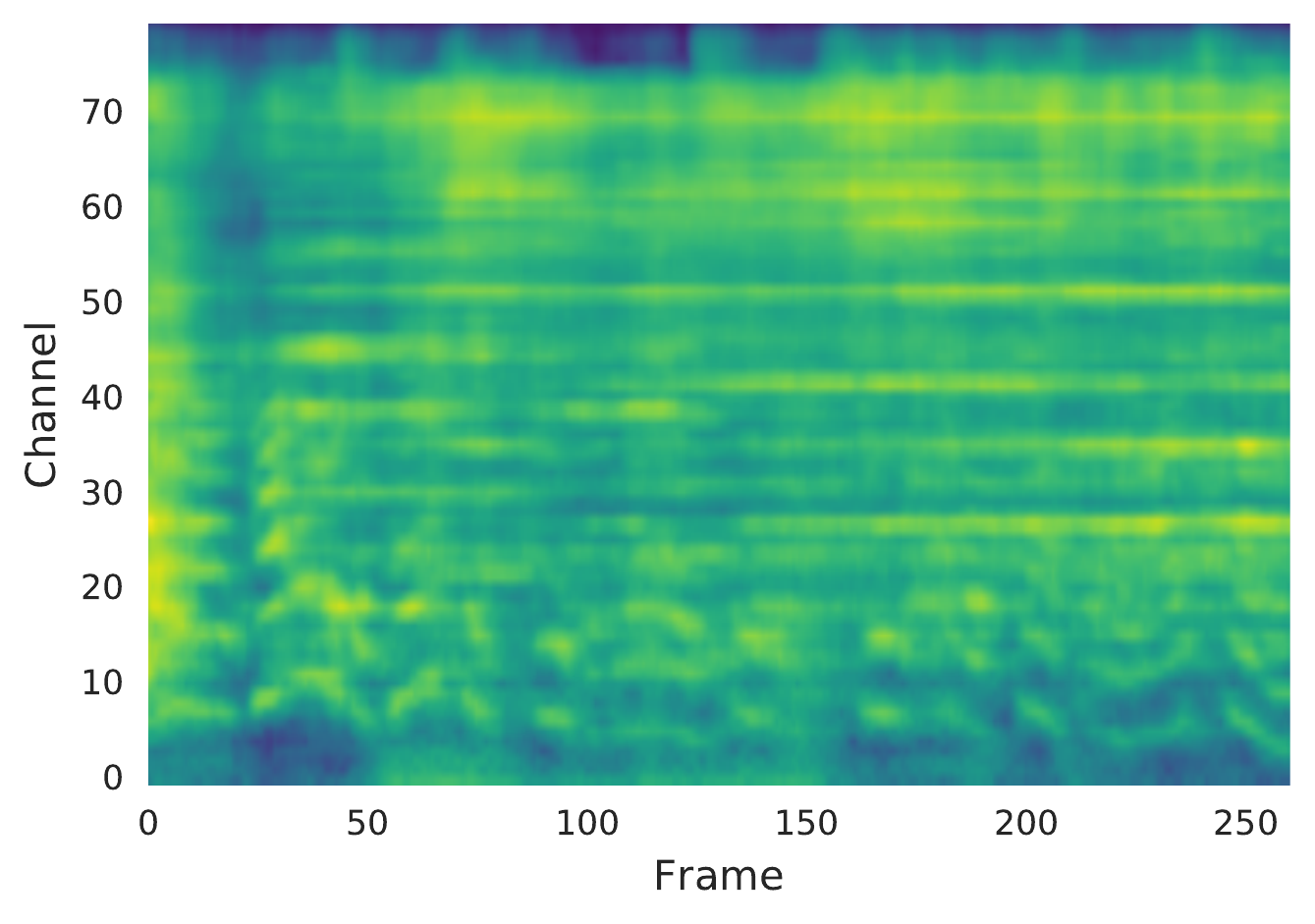}}
\subfigure[``Reverb.'' token] {\includegraphics[scale=0.28]{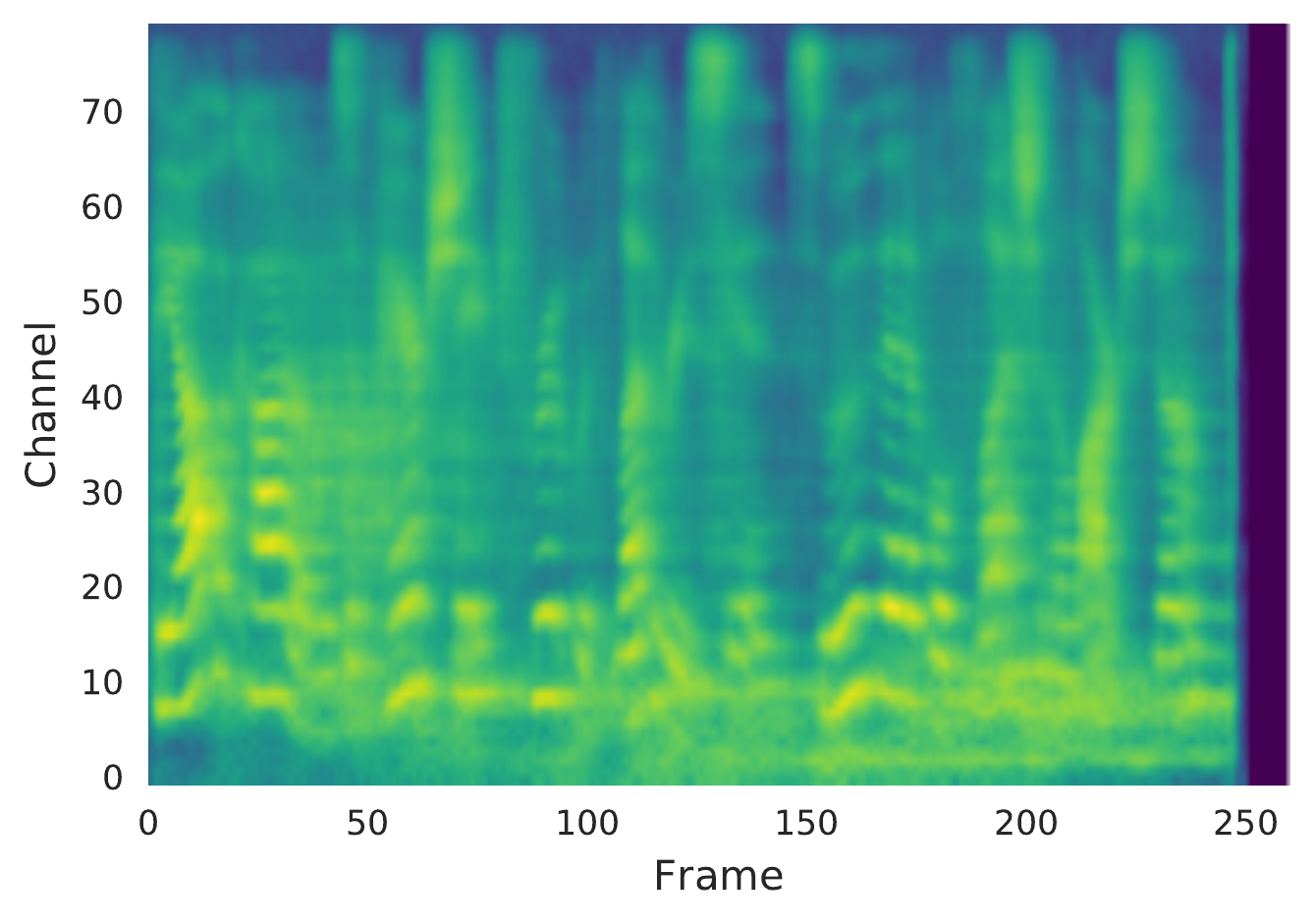}}
\subfigure[``Noise'' token] {\includegraphics[scale=0.28]{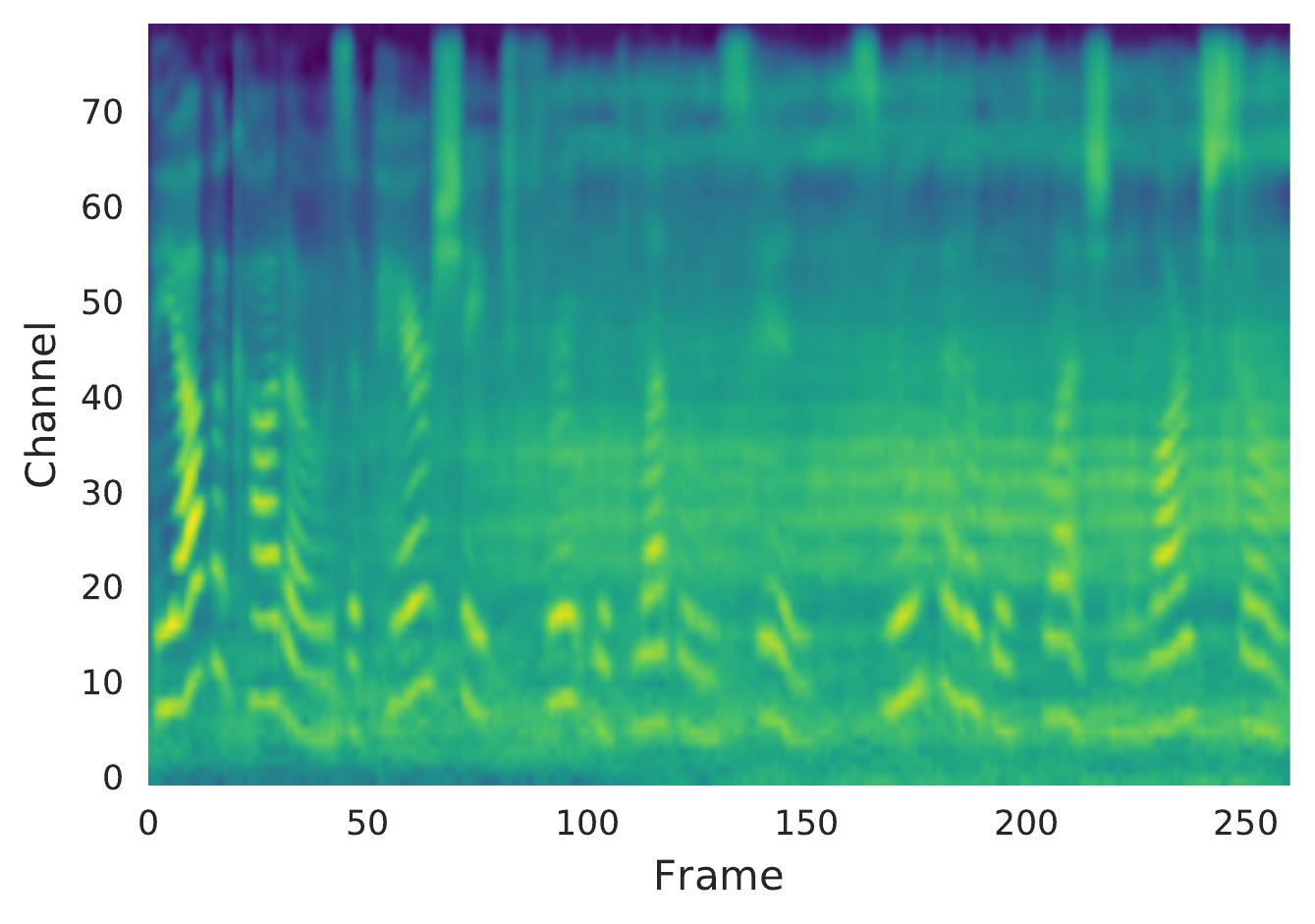}}
\subfigure[Clean token] {\includegraphics[scale=0.28]{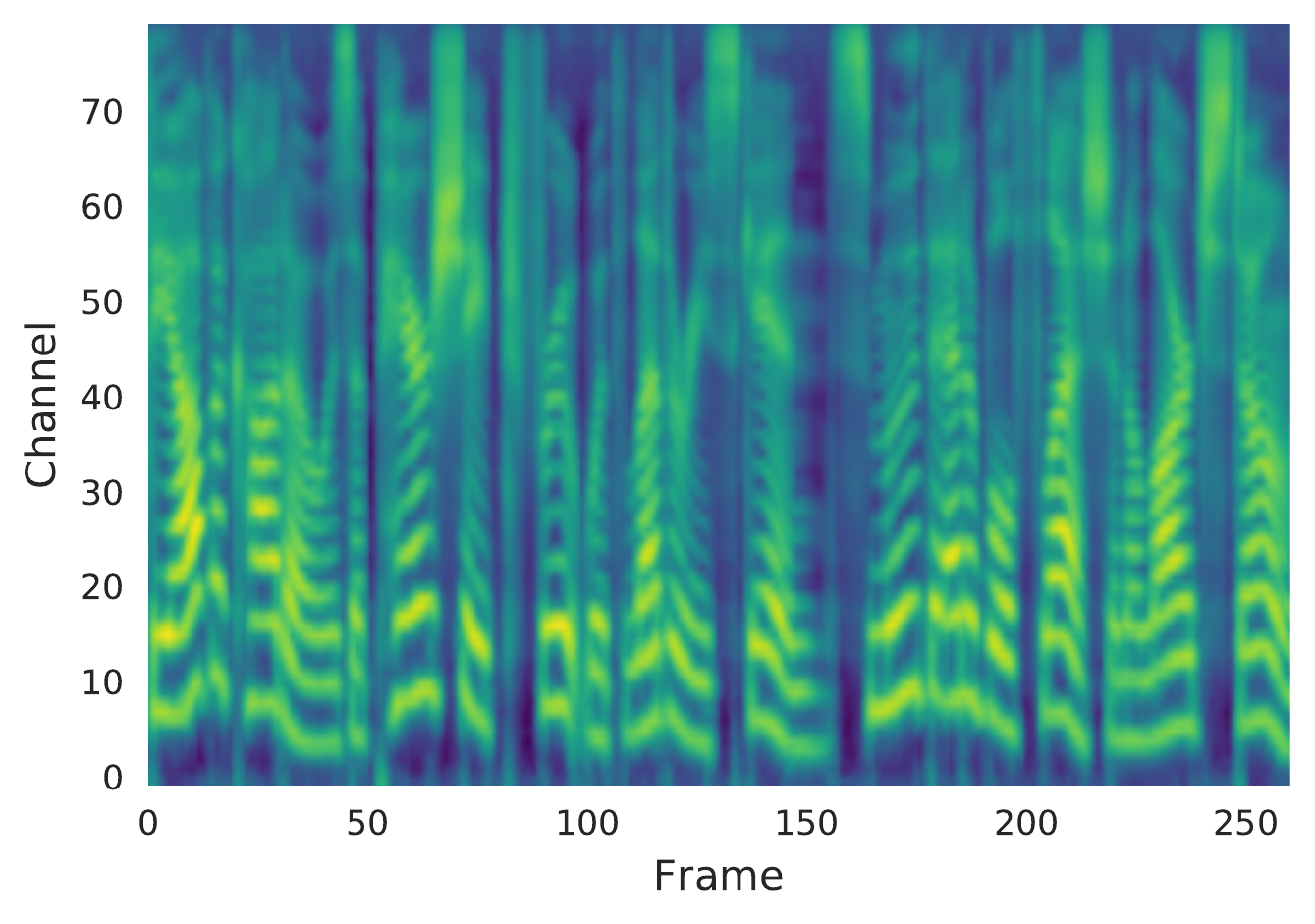}}
\caption{{\it Noisy and clean tokens uncovered.}}
\label{fig.expts.mtrtokens}
\vskip -0.1in
\end{figure}

After training a GST-augmented Tacotron on these datasets, we run inference in the first mode described in Section \ref{sec.model.inference}. Instead of providing a reference signal, we condition the model on each individual style token, which gives us an interpretable, audible sense of what each token has learned.
Interestingly, we find that different noises are treated as styles and ``absorbed'' into 
different tokens. We illustrate the spectrograms from a few tokens in 
Figure \ref{fig.expts.mtrtokens}. We can see (and
 \href{https://google.github.io/tacotron/publications/global_style_tokens/index.html#noisy_data.learned_tokens}{hear}) 
that these tokens clearly correspond to different interference types, such as music, reverberation and general background noise. Importantly, this method reveals that a subset of the learned tokens also correspond to completely clean speech. This means that we can synthesize clean speech for arbitrary text input by conditioning the model on a single, clean style token.

To demonstrate this, we run inference using a manually-identified clean style token (scaled to 0.3), and then evaluate the output using MOS naturalness tests.
We use the same 100-phrase evaluation set as \cite{yx2017tacotron}, collecting 8 ratings each from crowdsourced native speakers.
Table \ref{tb.mtr.mos} shows MOS results for both a baseline Tacotron and a ``clean-token" GST 
model. While the baseline Tacotron achieves a 4.0 MOS when the dataset is 100\% clean, 
MOS decreases as interference increases, dropping to a low score of 1.353. 
Because the model has no prior knowledge of speech or noise, it blindly models all statistics in the training set, resulting in substantial amounts of noise during synthesis.

By contrast, the GST model achieves about 4.0 MOS in all noise conditions. Note that the number of tokens needs to increase along with the percentage of noise to achieve this result. For example, a 10-token GST model yields clean tokens when trained on a 50\% noise dataset, but the noisier datasets required a 20-token model. Future work may explore how to adapt the number of tokens automatically to a given data distribution.

Audio examples from these models can be found \href{https://google.github.io/tacotron/publications/global_style_tokens/index.html#noisy_data}{here}.

\subsection{Real Multi-Speaker Found Data}

\begin{figure}[t]
\centering
\subfigure[Token A] {\includegraphics[scale=0.3]{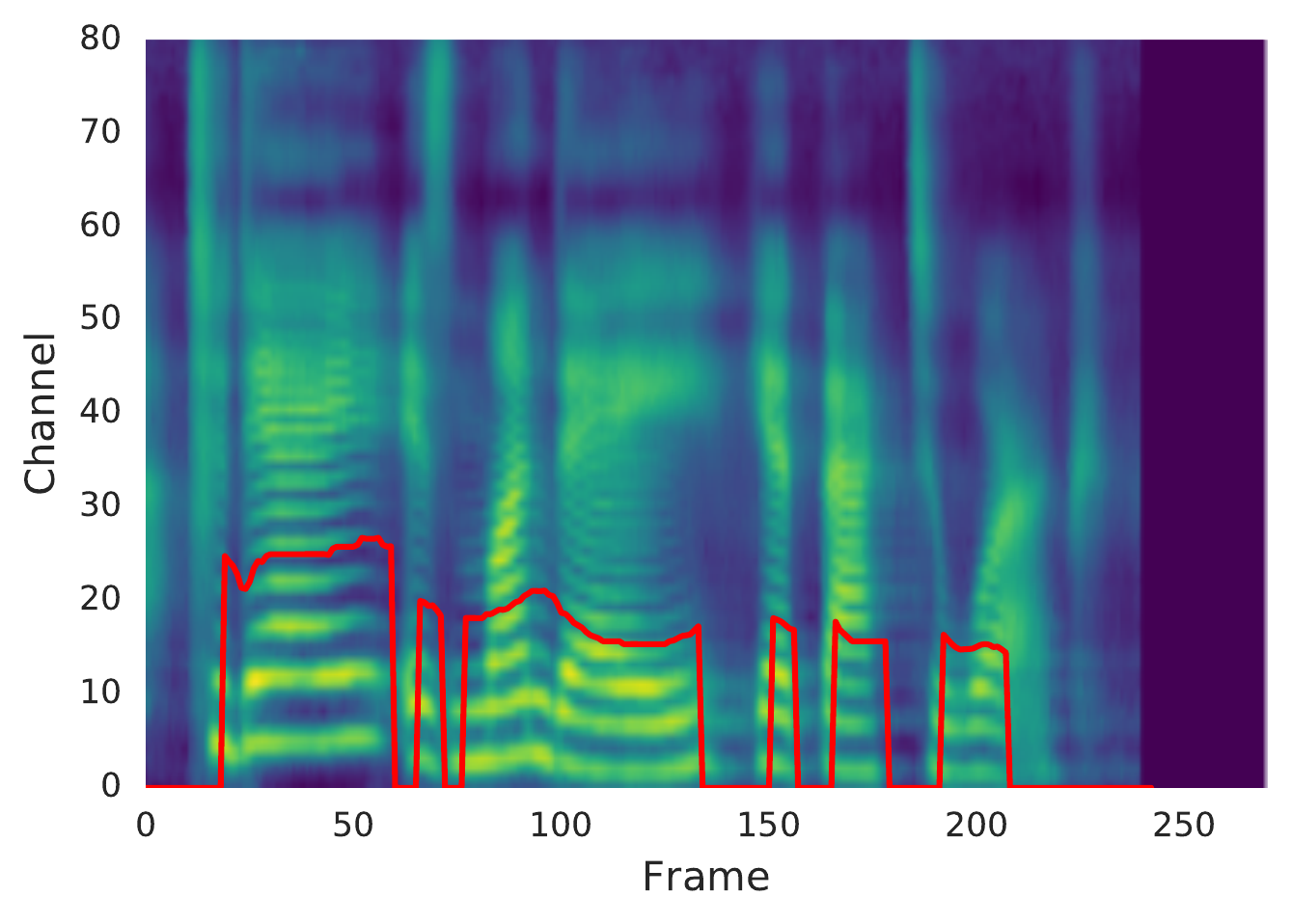}}
\subfigure[Token B] {\includegraphics[scale=0.3]{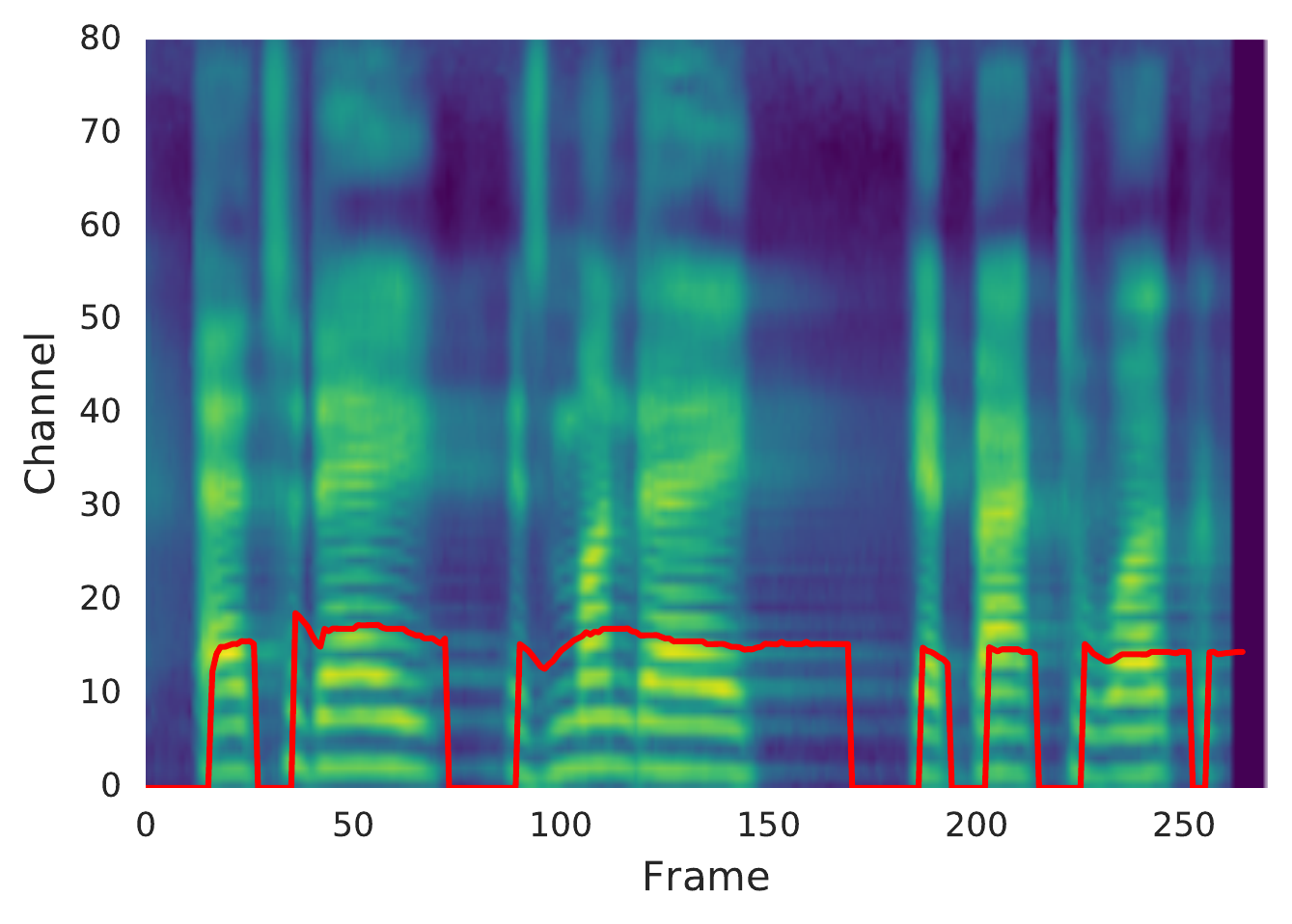}}
\caption{{\it Log-mel spectrograms (overlaid with F0 tracks) of two randomly chosen tokens from a GST model trained on the TED data. The two tokens uncover two different speakers.}}
\label{fig.ted.tokens}
\end{figure}

Our second experiment uses real data. This dataset is made up of audio tracks mined from 439 official TED YouTube channel videos. The tracks contain significant acoustic variations, including channel variation (near- and far-field speech), noise (e.g. laughs), and reverberation. We use an endpointer to segment the audio tracks into short clips, followed by an ASR model to create $<$text, audio$>$ training pairs.
Despite the fact that the ASR model generates a significant number of transcription and misalignment errors, we perform no other preprocessing. The final training set is about 68 hours long and contains about 439 speakers.

Without using any metadata as labels, we train a baseline Tacotron and a 1024-token GST model for comparison. As expected, the baseline fails to learn, since the multi-speaker data is too varied. The GST model results are presented in Figure \ref{fig.ted.tokens}. This shows spectrograms for the same phrase overlaid with F0 tracks, generated by conditioning the model on two randomly chosen tokens. Examining the trained GSTs, we find that different tokens correspond to different speakers. 
This means that, to synthesize with a specific speaker's voice, we can simply feed audio from that speaker as a reference signal. See Section \ref{sec.ted.quant} for more quantitative evaluations.

\begin{table}[t]
\caption{WER for the Spanish to English unsupervised language transfer experiment. Note that WER is an underestimate of the true intelligibility score; we only care about the relative differences.}
\label{tb.ted.wer}
\vskip 0.15in
\begin{center}
\begin{sc}
\begin{tabular}{cc}
\toprule
Model & WER (ins/del/sub) \\
\midrule
GST    & 18.68 (6.13/2.37/10.18) \\
Multi-speaker & 56.18 (3.75/20.27/32.14) \\
\bottomrule
\end{tabular}
\end{sc}
\end{center}
\vskip -0.2in
\end{table}

Finally, we exploit the fact that most of the talks are in English, but a small fraction are in Spanish. For this experiment, we compare baseline and GST-enabled noisy data models on a cross-lingual style transfer task. For a baseline, we train a multi-speaker Tacotron similar to \cite{ping2017deep}, using video IDs as a proxy for speaker labels. Conditioned on a Spanish speaker label, we then synthesize 100 English phrases.
For the GST system, we feed a reference signal from the same Spanish speaker and synthesize the same 100 English phrases. 
While the Spanish accent from the speaker is not preserved, we find that the GST model produces completely intelligible English speech with a similar pitch range as the speaker. 
By contrast, the multi-speaker Tacotron output is much less intelligible. 

To evaluate this result objectively, we compute word error rates (WER) of an English ASR model on the synthesized speech.
As shown in Table \ref{tb.ted.wer}, the WER of the GST utterances is much lower than that of the multi-speaker model.

The results strongly corroborate that GSTs learn embeddings disentangled from text content. Though this is an exciting early result, an in-depth study of using GST for prosody-preserving language transfer is in order.

\subsection{Quantitative Evaluations}
\label{sec.ted.quant}

\begin{figure*}[t]
\centering
\subfigure[50\% noisy data] {\label{fig.ted.SNE1}\includegraphics[scale=0.33]{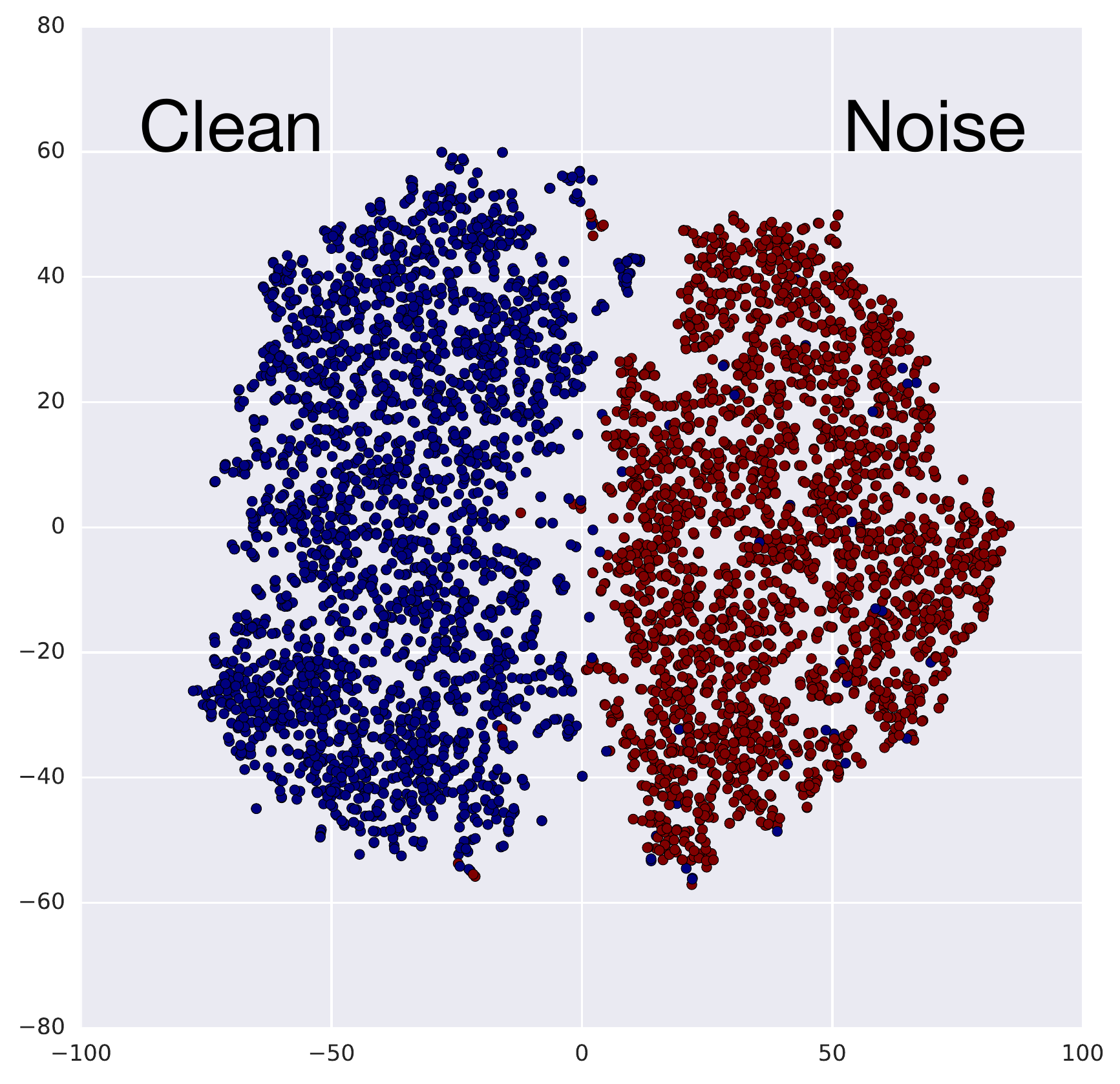}}
\hfil
\subfigure[Multi-speaker TED] {\label{fig.ted.SNE2}\includegraphics[scale=0.33]{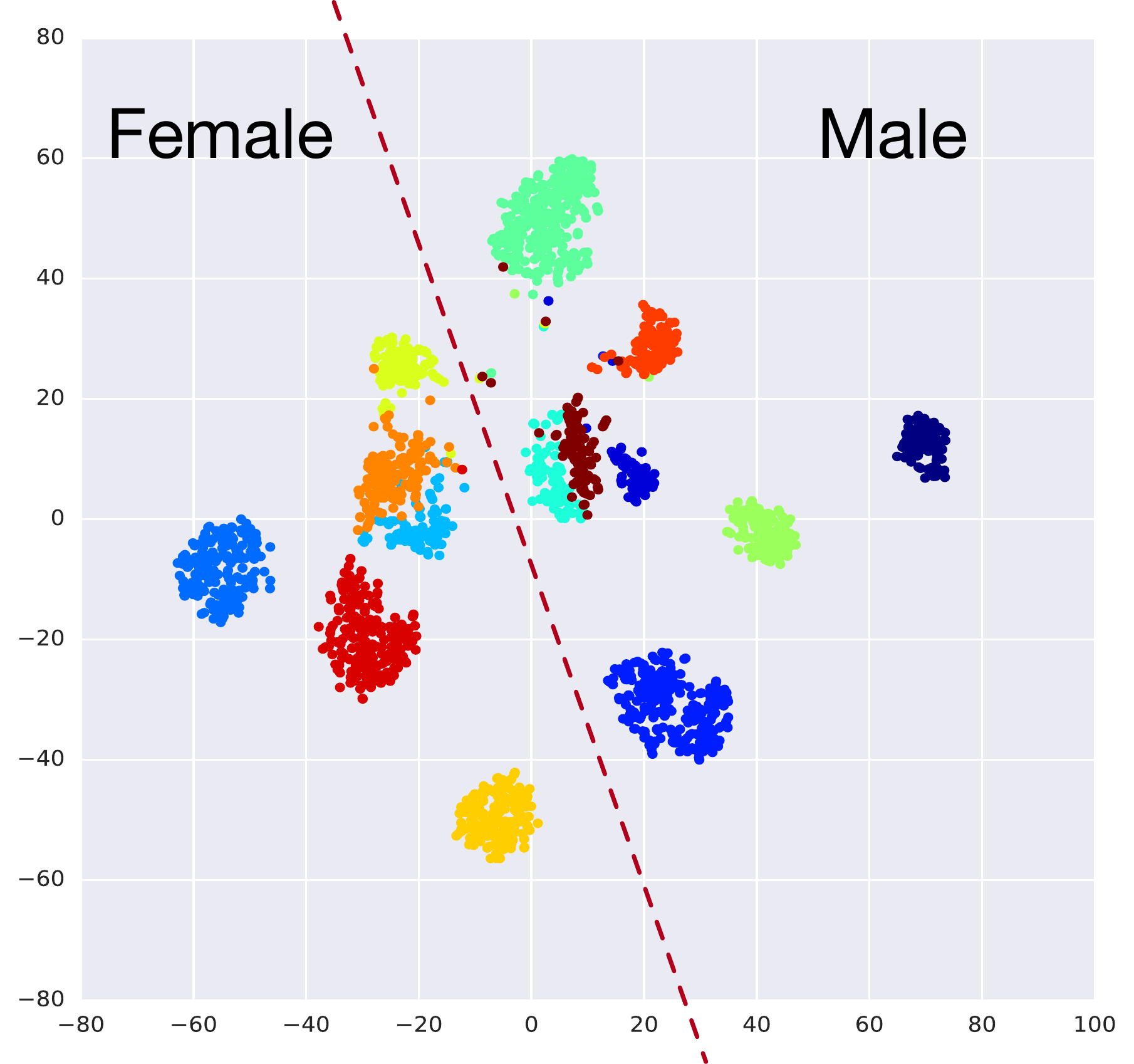}}
\caption{{\it Style embedding visualization using t-SNE.}}
\vskip -0.1in
\end{figure*}

We use t-SNE \cite{maaten2008visualizing} to visualize the style embeddings learned from both the artificial noise and TED datasets. Figure \ref{fig.ted.SNE1} shows that the embeddings learned from the artificial noisy dataset (50\% clean) are clearly separated into two classes. Figure \ref{fig.ted.SNE2} shows style embeddings for 2,000 randomly drawn samples containing 14 TED talk data speakers. We see that samples are well separated into 14 clusters, each corresponding to an individual speaker. Female and male speakers are linearly separable.

We also use style embeddings as features to perform noise and speaker classification with Linear Discriminative Analysis. 
Results are shown in Table \ref{tb.ted.spkacc}. For noise classification, GSTs uncover the true label with 99.2\% accuracy. For speaker classification, we use TED video IDs as true labels and compare with the $i$-vector method \cite{dehak2011front}, a standard representation used in modern speaker verification systems. For this task, the test set contains 431 speakers. While both trained and tested on short utterances (mean duration 3.75 secs), we can see that GSTs are comparable with $i$-vectors. This is an encouraging result, given that $i$-vectors were specifically designed for speaker classification. We speculate that GST has the potential to be applied to speaker diarization.

\subsection{Implications}
The results above have important implications for future TTS research on found data. First, due to the robustness of GSTs to both acoustic and textual noise, the design of automated data mining pipelines may be greatly simplified. Accurate segmentation and ASR models, for example, are no longer necessary to build high-quality TTS models. Second, style attributes, such as emotion, are often very difficult to label for large-scale noisy data. Using GSTs or weights to automatically generate style annotations may substantially reduce the human-in-the-loop efforts.

\begin{table}[t]
\caption{Classification accuracy (noise-vs-clean and TED speaker ID) using GSTs and $i$-vectors. Despite being trained within a generative model, GSTs encode rich discriminative information.}
\label{tb.ted.spkacc}
\vskip 0.15in
\begin{center}
\begin{small}
\begin{sc}
\begin{tabular}{ccc}
\toprule
Embedding & Artificial Data & Ted (431 speakers)  \\
\midrule
GST    & 99.2\% & 75.0\%  \\
i-vector & / &  73.4\%\\
\bottomrule
\end{tabular}
\end{sc}
\end{small}
\end{center}
\end{table}

%% file: 8-conclusions.tex
\section{Conclusions and Discussions}
\label{sec.conclusion}

This work has introduced Global Style Tokens, a powerful method for modeling style in end-to-end TTS systems.
GSTs are intuitive, easy to implement, and learn without explicit labels.
We have shown that, when trained on expressive speech data, a GST model yields interpretable embeddings that can be used to control and transfer style. 
We have also demonstrated that, while originally conceived to model speaking styles, GSTs are a general technique for uncovering latent variations in data. 
This was corroborated by experiments on unlabeled noisy found data, which showed that the GST model learns to decompose various noise and speaker factors into separate style tokens.

There is still much to be investigated, including improving the learning of GST, and using GST weights as targets to predict from text. Finally, while we only applied GST to Tacotron in this work, we believe it can be readily used by other types of end-to-end TTS models. More generally, we envision that GST can be applied to other problem domains that benefit from interpretability, controllability and robustness. For example, GST may be similarly employed in text-to-image and neural machine translation models.